\documentclass[onefignum,onetabnum]{siamonline220329}

\usepackage{lipsum}
\usepackage{amsfonts}
\usepackage{graphicx}
\usepackage{epstopdf}
\usepackage{algorithmic}
\ifpdf
  \DeclareGraphicsExtensions{.eps,.pdf,.png,.jpg}
\else
  \DeclareGraphicsExtensions{.eps}
\fi

\usepackage{enumitem}
\setlist[enumerate]{leftmargin=.5in}
\setlist[itemize]{leftmargin=.5in}

\newsiamremark{remark}{Remark}
\newsiamremark{hypothesis}{Hypothesis}
\crefname{hypothesis}{Hypothesis}{Hypotheses}
\newsiamthm{claim}{Claim}

\headers{Symmetrical MMD OT-Flow}{Zhe Xiong, Qiaoqiao Ding, and Xiaoqun Zhang}

\title{Arbitrary Distributions Mapping via SyMOT-Flow: A Flow-based Approach Integrating Maximum Mean Discrepancy and Optimal Transport}

\author{Zhe Xiong\thanks{School of Mathematical Sciences, Shanghai Jiao Tong University, 200240 Shanghai, China.
  (\email{aristotle-x@sjtu.edu.cn}).}
\and Qiaoqiao Ding\thanks{Institute of Natural Sciences,  School of Mathematical Sciences, MOE-LSC \& Shanghai National Center for Applied Mathematics (SJTU Center), Shanghai Jiao Tong University, 200240 Shanghai, China.
(\email{dingqiaoqiao@sjtu.edu.cn}, \email{xqzhang@sjtu.edu.cn}).}
\and Xiaoqun Zhang\footnotemark[3]}

\usepackage{amsopn}

\usepackage{bm, mathrsfs, amssymb,amsmath,graphicx,subfigure,color}
\usepackage{float}
\usepackage{multicol, multirow}
\usepackage{ulem} 
\usepackage[utf8]{inputenc}

\usepackage{booktabs}
\usepackage{makecell}
\usepackage{arydshln}

\usepackage{array}
\usepackage{arydshln}

\newcommand{\R}{\mathbb{R}}
\newcommand{\E}{\mathbb{E}}

\newcommand{\spaceH}{\mathcal{H}}
\newcommand{\spaceM}{\mathcal{M}}
\newcommand{\spaceT}{\mathcal{T}}
\newcommand{\datax}{\mathbf{x}}
\newcommand{\datay}{\mathbf{y}}
\newcommand{\dataz}{\mathbf{z}}

\newcommand{\dataxp}{\mathbf{x}^\prime}

\newcommand{\datazp}{\mathbf{z}^\prime}
\newcommand{\boldone}{\mathbf{1}}
\newcommand{\Np}{N^\prime}
\newcommand{\mmd}{\mathrm{MMD}}
\newcommand{\ot}{\mathrm{OT}}
\newcommand{\invT}{T^{-1}}

\newtheorem{assumption}{Assumption}[section]

\ifpdf
\hypersetup{
  pdftitle={Arbitrary Distributions Mapping via SyMOT-Flow: A Flow-based Approach Integrating Maximum Mean Discrepancy and Optimal Transport},
  pdfauthor={Zhe Xiong,
        Qiaoqiao Ding,
        and Xiaoqun Zhang}
}
\fi

\begin{document}

\maketitle

\begin{abstract}
Finding a transformation between two unknown probability distributions from finite samples is crucial for modeling complex data distributions and performing tasks such as sample generation, domain adaptation and statistical inference. One powerful framework for such transformations is normalizing flow, which transforms an unknown distribution into a standard normal distribution using an invertible network. In this paper, we introduce a novel model called SyMOT-Flow that trains an invertible transformation by minimizing the symmetric maximum mean discrepancy between samples from two unknown distributions, and  an optimal transport cost is incorporated as regularization to obtain a short-distance  and interpretable transformation. The resulted transformation leads to more stable and accurate sample generation. Several theoretical results are established for the proposed model and  its effectiveness  is validated with low-dimensional illustrative examples as well as high-dimensional bi-modality medical image generation  through the forward and reverse flows.
\end{abstract}

\begin{keywords}
Optimal Transport, Normalizing Flow, Maximum Mean Discrepancy.
\end{keywords}

\begin{MSCcodes}
68U10, 94A08
\end{MSCcodes}

\maketitle

\section{Introduction}\label{sec:introduction}
Finding a transformation between two unknown probability distributions from samples has many applications in machine learning and statistics, for example density estimation \cite{criminisi2012decision} and  sample generation \cite{creswell2018generative,ruthotto2021introduction}, for both  we can use the transformation to generate new samples from the target distribution. Furthermore, finding a transformation between two unknown probability distributions can also be useful in domains such as computer vision, speech recognition, and natural language processing, where we often encounter complex data distributions. For example, in computer vision, we can use the transformation to model the distribution of images and generate new images with desired characteristics \cite{farahani2021brief,croitoru2023diffusion}.

There are several common techniques for finding the transformation between two probability distributions. Invertible neural network (INN) is a popular and powerful modeling technique whose architectures are invertible by  design, which has attracted significant attention in statics \cite{louizos2017multiplicative,papamakarios2021normalizing} machine learning fields \cite{kingma2018glow,zhou2019density, dinh2016density}. Practically, INNs tend to be realized by the composition of a series of special invertible layers called flow layers, mainly including coupling flows \cite{ dinh2016density, kingma2018glow} and neural ODEs\cite{chen2018neural,grathwohl2018ffjord}, where each layers is designed to be easy to compute and invert. By applying a sequence of such transformations to a simple distribution, such as a Gaussian distribution, one can generate more complex distributions that can be used to model complex datasets. On this purpose, the structure of INN tends to be elaborately designed such that the transformation is invertible and the Jacobian determinant is tractable.   And as a widely used generative model, it has a good performance for both sampling and density evaluation tasks  \cite{rezende2015variational,papamakarios2021normalizing}.

On the other hand, optimal transport (OT) \cite{monge1781memoire,kantorovich2006problem} is a classical mathematical framework involving finding the optimal mapping between two probability distributions that minimizes a cost function, such as the Wasserstein distance \cite{ruschendorf1985wasserstein,villani2009wasserstein}. Optimal transport has been applied to many different generative models and improves the quality and stability of the generated samples \cite{coeurdoux2023learning,coeurdoux2022sliced,gulrajani2017improved,arjovsky2017wasserstein,dai2020sliced}.

The integration of deep learning with OT  has resulted in significant advancements in learning the optimal plan between two sets of samples. Due to the complexity of managing push-forward constraints in Monge's and Kantorovich's problems, some approaches prioritize solving the dual or dynamical formulations of OT. For instance, Onken et al. \cite{onken2021ot}  proposed combining OT with normalizing flows\cite{dinh2016density, kingma2018glow} by utilizing Neural ODE \cite{chen2018neural} to approximate the transformation between the given data and the standard Gaussian distribution through an 
$L_2$ constraint on the velocity field in dynamical OT. Morel et al. \cite{morel2022turning} introduced a geometric method to achieve the optimal map of a given normalizing flow without imposing constraints on the architecture or the training procedure. Additionally, Korotin \cite{korotin2022neural}   addressed the weak OT formulation to construct a transport mapping between the source and target distributions in dual space, demonstrating impressive performance.

Alternatively, utilizing a practical metric between two probability distributions offers a viable approach to addressing the optimal transport (OT) constraints. The selection of these metric distances plays a crucial role in determining the performance and characteristics of generative models. A commonly used metric is the Kullback-Leibler (KL) divergence, which has been extensively applied in various generative models, including Generative Adversarial Networks (GANs) and Normalizing Flows (NFs) \cite{goodfellow2020generative,kingma2013auto,kobyzev2020normalizing}. However, in general OT problems where only samples are available, the KL divergence is unsuitable due to the absence of density functions. Consequently, alternative distance metrics, such as the Kernel Stein Discrepancy (KSD) \cite{liu2016kernelized,gorham2017measuring}, have been investigated for posterior approximation in generative models. These metrics provide new opportunities for enhancing the accuracy and efficiency of generative models in diverse applications \cite{hu2018stein,fisher2021measure}. Another significant metric is the Maximum Mean Discrepancy (MMD) \cite{gretton2012kernel}, which measures the difference in mean values between samples using a continuous function. For instance, \cite{manupriya2020mmd} proposed employing MMD regularization to relax and solve the corresponding discrete unbalanced OT problem. Similarly, in \cite{arbel2019maximum}, the authors presented a method for conducting the gradient flow of MMD and elucidated its relationship with OT.

Motivated by invertible transformation constructed in normalizing flow approaches, in this paper, we propose a method to learn an invertible transformation between two unknown distributions based on  given samples, namely \textbf{SyM}metrical \textbf{M}MD \textbf{OT}-\textbf{Flow} (SyMOT-Flow).
In this manuscript, we build upon our foundational research initially presented at a specialized workshop \cite{xiong2023symot}. Our extension involves the augmentation of our model to address higher-dimensional imaging tasks, specifically focusing on the generation of images between two distinct magnetic resonance imaging (MRI) modalities, namely T1 and T2. This advancement represents a significant leap in the application of our proposed model, demonstrating its adaptability and efficacy in more complex imaging scenarios.
Additionally, this study places a heightened emphasis on theoretical results, providing a detailed exploration of the convergence properties of our model. We rigorously demonstrate our model's alignment with the optimal transport problem, offering insights into its theoretical robustness and potential applications in complex scenarios. In our model, the two-direction maximum mean discrepancy (MMD) \cite{gretton2012kernel} is used to measure the discrepancy between the transformed samples to the original ones. Besides, we consider the OT cost in Monge's problem \cite{monge1781memoire} as a regularization.
Focusing on the convergence properties within the Reproducing Kernel Hilbert Space (RKHS), as detailed in Reference \cite{simon2020metrizing}, we establish the connection of the proposed model and Monge's problem. The application of $\Gamma$-convergence theory provides substantial evidence for the gradual alignment of the model's minimizer with that of the optimal transport plan. Further, inspired by theoretical frameworks in  \cite{kong2021universal,neumayer2021optimal} , the feasibility and existence of the invertible neural network, which is a critical component of the model, are substantiated.
The proposed model takes the advantages of kernel in MMD for capturing intrinsic structure of samples and the regularity and stability of parameterized optimal transport. And the transformation is constructed through a sequence of invertible network structure which enables continuity and invertibility between two distributions in high dimension. 
Together with encoder-decoder, the learned transformation in feature spaces facilitates an optimal correspondence between samples, a property that holds significant potential for various applications. These include generative modeling, feature matching, and domain adaptation, where the precise alignment of features is crucial for enhancing model performance and adaptability.  Extensive experiments on both low-dimension illustrative examples and high dimension datasets demonstrate the performance of our model. Also, ablation studies on the effect of the OT regularization and symmetrical designs of our models are provided to show the characteristic of learned transformation. 

Unlike most approaches that solve Optimal Transport (OT) using neural networks in the dual space, our method focuses on the original space, allowing us to use a single network to accurately approximate the optimal mapping. In contrast, other methods such as those by Korotin et al. \cite{korotin2022neural} and Gushchin et al.  \cite{gushchin2024entropic}  require additional networks to learn the corresponding potential functions, thereby increasing model complexity and training difficulty. Additionally, Manupriya et al.\cite{manupriya2020mmd}  utilized Maximum Mean Discrepancy (MMD) as a distribution metric to solve the unbalanced OT problem. However, their practical method converts the MMD regularization into a matrix normalization to solve the corresponding discrete problem, primarily targeting low-dimensional tasks. In our model, we incorporate MMD into the training loss to measure the distance between the source and target distributions. This approach enables us to directly obtain an invertible continuous mapping between the source and target samples, making it suitable for high-dimensional image transfer tasks, such as MRI T1/T2 and CT/MRI, achieving generation results with high precision as demonstrated in our experiments.

This paper is organized as follows.
Section \ref{bg} gives the Definitions and preliminary Lemmas.
Section  \ref{method} describes the proposed method and Section \ref{theoretical results} gives the theoretical results.
Section \ref{experiments} is devoted to the experimental evaluation and comparison to other methods.
Section \ref{Conclusion} concludes the paper.

\section{Definitions and Lemmas}\label{bg}

Before introducing our model, we first introduce some notations and definitions that will be used for later theoretical analysis.

Suppose $\Omega \subset \R^d$ is a compact set where $d$ is the dimension and $\datax$, $\dataz$ are two random variables in $\Omega$ with distribution $p$, $q$. In practice, we only have some samples from $p$ and $q$, which are denoted by $\{\datax_i\}_{i=1}^N$ and $\{\dataz_j\}_{j=1}^{N^\prime}$ respectively, where $N$ and $N^\prime$ are the numbers of samples. Then correspondingly, we have the following definitions:
\begin{definition}[Reproducing Kernel Hilbert Space (RKHS)]
    Let $\spaceH$ be a Hilbert space of real-value functions on $\Omega$. A function $k: \Omega \times \Omega \to \R^d$ is a \textit{reproducing kernel of $\spaceH$} if $k(\cdot, \datax) \in \spaceH$ for all $\datax \in \Omega$, and $\langle f, k(\cdot, \datax)\rangle_\spaceH = f(\datax)$ for all $f \in \spaceH, \datax\in \Omega$. A Hilbert space $\spaceH$ with kernel $k(\cdot, \cdot)$ is called an RKHS.
\end{definition}

\begin{definition}[Strictly Integrally Positive Definite Kernel ($\int$s.p.d)]
    Suppose $\spaceM(\Omega)$ as the space of all finite signed measures on $\Omega$ and for $\mu \in \spaceM(\Omega)$, let $\mathcal{L}_1(\mu)$ be the space of all measurable functions $f:\Omega \to \R$ which are $\mu$-integrable. Suppose $\mathcal{H}$ is an RKHS with a $\mu$-measurable kernel $k$ and we define $\spaceM_\spaceH(\Omega)$ as follows:
    \begin{equation*}
        \spaceM_\spaceH(\Omega) := \left\{\mu\in \spaceM(\Omega) | \spaceH \subseteq \mathcal{L}_1(\mu)\right\}.
    \end{equation*}
    Then the kernel $k$ of $\spaceH$ is called strictly integrally positive definite ($\int$s.p.d) if the following condition always holds:
    \begin{equation*}
        \int_\Omega \int_\Omega k(\datax, \datax^\prime) d\mu(\datax) d\mu(\datax^\prime) \geq 0
    \end{equation*}
    for all the measures $\mu \in \spaceM_\spaceH(\Omega)$ and the equality holds if and only if $\mu \equiv 0$.
\end{definition}

\begin{definition}[Universal]\label{def:universal}
    An RKHS $\spaceH$ with kernel $k$ defined on a compact set $\Omega$ is called universal, if $\spaceH$ is dense in $\mathcal{C}_0(\Omega)$ with respect to $L_\infty$-norm and $k$ is continuous, where $\mathcal{C}_0(\Omega)$ denotes the space of continuous functions vanishing at boundary of $\Omega$.
\end{definition}

\begin{definition}[Diffeomorphism]\label{def:diffeomorphism}
    A differentiable map $T$ defined on $\Omega$ is called a ($C^{\infty}$-) diffeomorphism if it is invertible and smooth, i.e. $T$ is a bijection and both $T$ and $T^{-1}$ are ($C^{\infty}$-) differentiable.
\end{definition}

\subsection{Maximum Mean Discrepancy (MMD)}
    Suppose $p$ and $q$ are two probability measures defined on $\Omega$ and $\spaceH$ is an RKHS with kernel $k$. Then the maximum mean discrepancy between $p$ and $q$ respect to $\spaceH$ is defined as
    \begin{equation}\label{eq:mmd1}
        \mmd(\spaceH, p, q) = \sup_{\|f\|_{\spaceH}\leq 1} \mathbb{E}_p \left [f(\datax)\right] - \mathbb{E}_q \left [f(\dataz)\right].
    \end{equation}
    Moreover, since $\spaceH$ is a Hilbert space, by the Riesz representation theorem, there exists a feature mapping $\phi(\datax) \in \spaceH$ such that $f(\datax) = \langle f, \phi(\datax) \rangle_\spaceH$ for $\forall \datax \in \Omega$. And according to \cite{steinwart2008support}, this feature mapping takes the canonical form $\phi(\datax) = k(\datax,\cdot)$. In particular, we have that $\langle\phi(\datax),\phi(\dataz)\rangle_\spaceH = k(\datax, \dataz)$.

One important property of MMD is that, when the RKHS $\spaceH$ is universal, the MMD can be regarded as a metric of the probability measures, which is stated as following Lemma:
\begin{lemma}\cite[Theorem 5]{gretton2012kernel}
    Suppose $\spaceH$ is a universal RKHS, then the MMD is a metric of the probability measures. More precisely, suppose $p$ and $q$ are two probability measures defined on $\Omega$, then we have $\mmd(\spaceH, p, q) = 0$ if and only if $p = q$.
\end{lemma}
While in practice, the operator of $\sup$ on the set $\|f\|_\spaceH \leq 1$ is intractable, we tend to use the equivalent formulation of MMD ,which is also given as follows:
\begin{lemma}\cite[Theorem 6]{gretton2012kernel}\label{mmd}
    Suppose $\datax$ and $\datax^\prime$ are two independent random variables with distribution $p$, $\dataz$ and $\dataz^\prime$ are two independent random variables with distribution $q$, then the squared $\mmd$ is given by
    \begin{alignat*}{2}\label{eq:mmd2}
        \mmd^2(\spaceH, p, q) & = \|\mathbb{E}_{\datax\sim p}\left[\phi(\datax)\right] - \mathbb{E}_{\dataz\sim q}\left[\phi(\dataz)\right]\|_\spaceH^2\\
        & = \mathbb{E}_{\datax\sim p, \dataxp \sim p}\left[k(\datax, \dataxp)\right] + \mathbb{E}_{\dataz\sim q, \datazp \sim q}\left[k(\dataz, \datazp)\right] - 2\mathbb{E}_{\datax\sim p, \dataz \sim q}\left[k(\datax, \dataz)\right].
    \end{alignat*}
    Empirically, for the samples $\{\datax_i\}_{i=1}^N$ and $\{\dataz_j\}_{j=1}^{\Np}$ from distribution $p$ and $q$ respectively, we have the discrete estimator of MMD:
\begin{equation}\label{eq:mmd3}
     \mmd_b^2(\spaceH, p, q)
     = \frac{1}{N \Np}\sum_{n=1}^{N}\sum_{n^\prime=1}^{\Np}   \left[k(\datax_n, \datax_{n^\prime})  + k(\dataz_n, \dataz_{n^\prime}) - 2k(\datax_n, \dataz_{n^\prime})\right].
\end{equation}
\end{lemma}
We note that by taking MMD as a metric, some equivalent relationships between different types of convergence are provided in \cite{simon2020metrizing}. We choose part of them as the statement in the following lemma, which will be used in our later demonstration:
\begin{lemma}\cite[Lemma 3]{simon2020metrizing}\label{lm1}
    Suppose RKHS $\spaceH$ is universal with a continuous and $\int$s.p.d kernel $k$. Let $p_\alpha$ (sequence) and $p$ be probability measures. Then strong convergence with respect to MMD is equivalent to the weak convergence from $p_\alpha$ to $p$. Precisely, 
    \begin{equation*}
        \mmd(\spaceH, p_\alpha, p) \to 0 \iff \mathbb{E}_{p_\alpha} f \to \mathbb{E}_{p} f \text{ for all $f \in \mathcal{C}_b(\Omega)$}.
    \end{equation*}
\end{lemma}

\begin{remark}
Note that this estimator is a biased one while there is also unbiased estimator of MMD. The biased estimator, owing to its convenience in computation and prevalent usage in practical scenarios, is the primary focus of our discussion. More discussion on unbiased estimators of MMD can be found in \cite{gretton2012kernel}.
\end{remark}

\begin{remark}
While the definition of Maximum Mean Discrepancy (MMD) is contingent on the selection of the Reproducing Kernel Hilbert Space (RKHS), denoted as $\spaceH$, in practical applications, the specific choice of 
$\spaceH$ is often not a primary concern.  For convenience, we simplify the notation $\mmd(\spaceH, p, q)$ to $\mmd(p, q)$ in later discussions as  a metric between $p$ and $q$.
\end{remark}

\subsection{Optimal Transport (OT)}
Suppose $c(\cdot,\cdot):\Omega\times\Omega\rightarrow \R_{+}$ is a nonnegative cost function, $p$ and $q$ are two probability measures, then the optimal transport problem by \cite{monge1781memoire} is given by
\begin{equation}\label{prob:monge}
    \min_T\int_\Omega c(\datax, T(\datax)) d p(\datax)\quad s.t.~ T_\sharp p = q,
\end{equation}
where $T:\Omega\rightarrow\Omega$ is a measurable mapping and $T_\sharp$ is the push-forward operator such that
\begin{equation*}
    \left[\right. T_\sharp  p  \left. = q\right] \Longleftrightarrow \left[  \int_{\Omega}h(\dataz)dq(\dataz) = \int_{\Omega}h(T(\datax))dp(\datax), \forall h \in \mathcal{C}_b(\Omega)\right].
\end{equation*}
As in practice it is difficult to enhance the constraints from sampled data, we consider a relaxed form of problem 
\eqref{prob:monge} by replacing the equality $T_\sharp p = q$ with a probability measure distance $d(\cdot,\cdot)$ as
\begin{equation}\label{prob:relaxed monge}
    \min_T \int_{\Omega} c(\datax, T(\datax))dp(\datax) + \lambda d(T_\sharp p, q),
\end{equation}
where $\lambda > 0$ is the weight of the distance penalty.

\section{Our method}\label{method}
\begin{figure*}[htbp]
    \centering
    \includegraphics[width=1.0\linewidth]{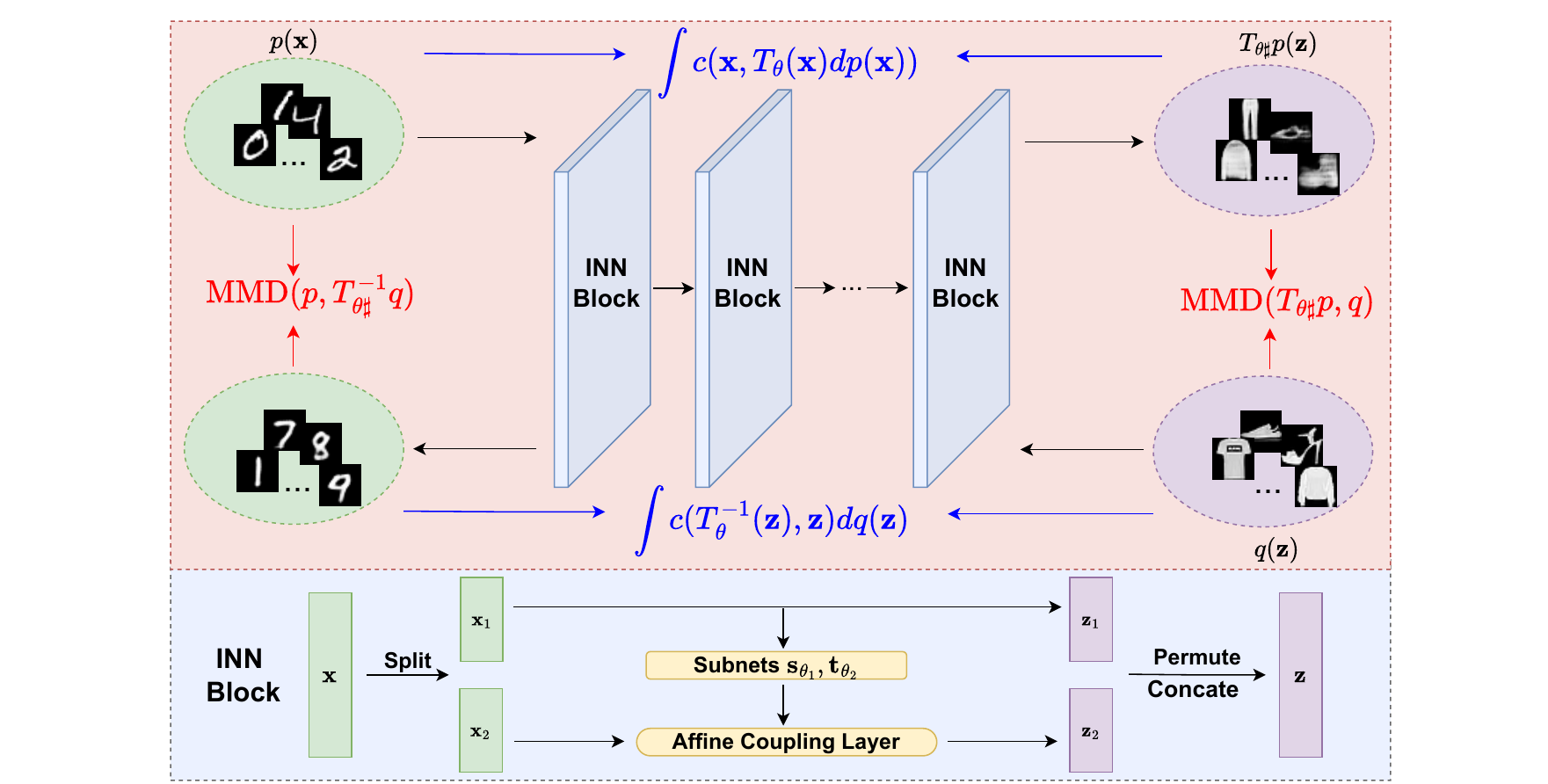}
    \caption{Overview of SyMOT-Flow Model. The \textbf{red} block contains the main structure of the model and the \textbf{blue} one consists of the precise structure of the INN blocks, which are the basic block of the distribution transformation $T_\theta$.}
    \label{fig:pipeline}
\end{figure*}
Our goal is to develop a methodology that identifies a transformation, parameterized by Invertible Neural Networks (INNs), to map between two distributions using sample data. The diagram of our model is presented in Fig.  \ref{fig:pipeline}. The red block contains the main process of the method and the blue one consists of the precise structure of the INN blocks, which are the basic block of the distribution transformation between the two given data distributions (the green set and purple set). In the following, we will present the details of the proposed method.

\subsection{Invertible Neural Networks (INNs)}
Invertible Neural Networks (INNs) are neural networks architectures with invertibility by design, which are often composed of invertible modules such as affine coupling layers \cite{dinh2016density} or neural ODE \cite{chen2018neural}. With these specially designed structures, it tends to be tractable to compute the inverse transformation and Jacobian determinant, which is widely used in the normalizing flow tasks. As we can see in Fig. \ref{fig:pipeline}, the input $\datax$ is split along the channels into two part $(\datax_1, \datax_2)$ randomly and then the first part $\datax_1$ keeps invariant, while the second part $\datax_2$ are fed into two trainable subnets $\mathbf{s}_{\theta_1}$ and $\mathbf{t}_{\theta_2}$. Furthermore, through the affine coupling layer, the output $\dataz_2$ is determined by the affine combination of $\mathbf{s}_{\theta_1}(\datax_1))$, $\mathbf{t}_{\theta_2}(\datax_1)$ and $\datax_2$. In the last step, the output $(\dataz_1, \dataz_2)$ are primarily concatenated and then reordered along the channels to get the transformed output $\dataz$. Here is the transformation in the affine coupling layer:
\begin{equation*}
    \begin{aligned}
        & \dataz_1 = \datax_1, \\
        & \dataz_2 = \datax_2 \odot \exp\left(\gamma * \tanh(\mathbf{s}_{\theta_1}(\datax_1))\right) + \mathbf{t}_{\theta_2}(\datax_1),
    \end{aligned}
\end{equation*}
where $\gamma$ is the affine clamp parameter. Here the subnets $\mathbf{s}_{\theta_1}(\cdot)$ and $\mathbf{t}_{\theta_2}(\cdot)$ can be designed differently, for example, the RealNVP \cite{dinh2016density} and Glow \cite{kingma2018glow}.
\subsection{Loss function}
As mentioned in Section \ref{bg}, we choose $d(\cdot,\cdot)$ to be the squared MMD in the relaxed OT \eqref{prob:relaxed monge}. Moreover, to improve the stability of the transformation $T$, we make use of the invertibility of $T$ and design a symmetrical distance as follows:
\begin{equation*}
    d_\mmd(T,p,q) = \mmd(T_\sharp p, q) + \mmd(p, {T^{-1}}_\sharp q).
\end{equation*}
Correspondingly, we also add a symmetrical cost to the objective function in OT and finally the loss function with parameter $\lambda$ is defined as
\begin{equation}\label{loss:original}
    L_T = \left(\int_{\Omega} c(\datax, T(\datax))d p(\datax) + \int_{\Omega} c(T^{-1}(\dataz), \dataz) d q(\dataz)\right)
     + \lambda d_\mmd(T,p,q).
\end{equation}
In practice, suppose $T_\theta$ is an invertible network with parameters $\theta$. Given two sets of samples $\{\datax_i\}_{i=1}^N$ and $\{\dataz_j\}_{j=1}^{N'}$. The empirical training loss function is defined as follows:
\begin{equation}\label{loss:empirical}
    \begin{aligned}
        L_\theta & = \frac{1}{NN'}\sum_{n=1}^N\sum_{n^\prime=1}^{N'} \left[ k(T_\theta(\datax_n), T_\theta(\datax_{n^\prime})) + k(T^{-1}_\theta(\dataz_n), T^{-1}_\theta(\dataz_{n^\prime})\right] \\
        & - \frac{2}{NN'}\sum_{n=1}^N\sum_{n^\prime=1}^{\Np} \left[k(T_\theta(\datax_n), \dataz_{n^\prime}) + k(\datax_n, T_\theta^{-1}(\dataz_{n^\prime}))\right] \\
        & + \frac{\beta}{N}\sum_{i=1}^{N} c(\datax_i, T_\theta(\datax_i)) + \frac{\beta}{\Np}\sum_{j=1}^{\Np} c(T_\theta^{-1}(\dataz_j), \dataz_j),
    \end{aligned}
\end{equation}
where $\beta = \frac{1}{\lambda} > 0$ is the weight parameter and $T_\theta^{-1}$ is the inversion of $T_\theta$. More precisely as shown in Fig. \ref{fig:pipeline}, $T_\theta$ is the composition of a series of INN blocks. In the forward direction, we use empirical $\mmd(T_{\theta\sharp} p, q)$ to ensure $\{T_\theta(\datax_n)\}$ obey the same distribution as $\{\dataz_n\}$. Moreover, the empirical OT objective $\int c(\datax, T_\theta(\datax))dp(\datax)$ is to obtain a transport map with as less cost as possible. And for the backward process, the corresponding inverse $\mmd(p, \invT_{\theta\sharp} q)$ and OT objective $\int c(\invT_\theta(\dataz),\dataz)dq(\dataz)$ also fulfill an analogous function.
\begin{remark}
Note that the empirical loss \eqref{loss:empirical} is slightly different from \eqref{loss:original} as we put the weight parameter on the  OT term  for actual implementation. As opposed to merely minimizing the OT cost, it is crucial to prioritizing the attainment of a close-to-zero MMD to establish the validity of the constraint $T_\sharp p = q$. Moreover, in the calculation of empirical MMD, we omit two items which are irrelevant to the parameters of the invertible network.
\end{remark}

\section{Theoretical Analysis}\label{theoretical results}
We introduce some theoretical results to explain the rationality and validation of our model. Before giving the theorems, we firstly make some assumptions as follows:
\begin{assumption}\label{assumption:N}
    The numbers of samples $N$ and $\Np$ from distributions $p$ and $q$ are both large enough to ensure that $P\left\{\lvert \mmd_b(p, q) - \mmd(p ,q)\rvert > \delta\right\} < \epsilon$ for small numbers $\delta$ and $\epsilon$, which implies that the empirical estimator $\mmd_b$ is closed to the true MMD such that the optimal transformation $T_\theta$ obtained from the samples is an accurate approximation to the continuous solution.
\end{assumption}

\begin{assumption}\label{assumption:T}
    The optimal transport problem \eqref{prob:monge} is solved in the space of all the diffeomorphisms defined on $\Omega$, which implies the existence of an optimal plan $T$ between the distribution $p$ and $q$ being invertible and smooth.
\end{assumption}
To be consistent with Assumption \ref{assumption:T}, we consider the symmetrical OT problem in our theorems. Recall the problem \eqref{prob:monge} and define the symmetrical version as $\ot(p, q)$:
\begin{equation}\label{prob:sym monge}
    \begin{aligned}
     & \min_T\int_{\Omega} c(\datax, T(\datax))d p(\datax) + \int_{\Omega} c(T^{-1}(\dataz), \dataz) d q(\dataz) \\
     & s.t.~~ T_\sharp p = q. 
\end{aligned}
\end{equation}
Correspondingly, the relaxed Monge's problem is defined as $\ot_\lambda(p, q)$:
\begin{equation}\label{prob:sym relax monge}
     \min_T \int_{\Omega} c(\datax, T(\datax))d p(\datax)  + \int_{\Omega} c(T^{-1}(\dataz), \dataz) d q(\dataz) + \lambda d_\mmd(T,p,q)
\end{equation}

The following theorem reveals the relationship between the optimal solutions of problem \eqref{prob:sym monge} and \eqref{prob:sym relax monge}:
\begin{theorem}\label{thm:relationship}
    Suppose $p$ and $q$ are two probability measures defined on $\Omega$ and $\datax$ and $\dataz$ two random variables that follow $p$ and $q$ distributions respectively. If the RKHS $\mathcal{H}$ is universal and its kernel function $k$ is $\int$s.p.d, then for any positive and increasing sequence $\left\{\lambda\right\}$,  it holds that,
    \begin{equation}
        \lim_{\lambda \rightarrow +\infty} \ot_\lambda(p, q) = \ot(p, q).
    \end{equation}
\end{theorem}
\begin{proof}[Proof of Theorem \ref{thm:relationship}]
    Suppose for each $\lambda > 0$, $T_\lambda^\star$ is an minimizer of problem $\ot_\lambda(p, q)$ and $T^\star$ is the minimizer of the original OT problem $\ot(p, q)$ respectively. By the definition of minimizer, for $T_\lambda^\star$ we have that
    \begin{equation*}
        \begin{aligned}
            &\int_{\Omega} c(\datax, T^\star(\datax))dp(\datax) + \int_{\Omega} c({T^\star}^{-1}(\dataz), \dataz) d q(\dataz) \\
         &\geq \int_{\Omega} c(\datax, T_\lambda^\star(\datax)) dp(\datax) + \int_{\Omega} c({T_\lambda^\star}^{-1}(\dataz), \dataz) d q(\dataz)
         + \lambda d_\mmd(T_\lambda^\star,p,q).
        \end{aligned}
    \end{equation*}
    Then consequently, it holds that
    \begin{equation}\label{eq:inequality1}
        \limsup_{\lambda \rightarrow +\infty}  \lambda d_\mmd(T_\lambda^\star,p,q) \leq \limsup_{\lambda \rightarrow +\infty} \ot_\lambda(p, q) \leq \ot(p, q) < +\infty.
    \end{equation}
    On the other hand, from inequality \eqref{eq:inequality1} it is easy to get that
    \begin{equation}\label{eq:lim}
        \lim_{\lambda \rightarrow +\infty} d_\mmd(T_\lambda^\star,p,q) = 0.
    \end{equation}
    According to Lemma \ref{lm1}, since the kernel function $k$ is $\int$s.p.d and $\mathcal{H} \subset \mathcal{C}_0$, the limit (\ref{eq:lim}) indicates that ${T_\lambda^\star}_\sharp p \rightarrow q$ and ${{T_\lambda^\star}_\sharp}^{-1}q\rightarrow p$ in weak sense. Hence we obtain 
    \begin{equation}\label{eq:inequality2}
        \liminf_{\lambda \rightarrow +\infty} \ot_\lambda(p, q)
        \geq \liminf_{\lambda \rightarrow +\infty}\left[ \int_{\Omega} c(\datax, T_\lambda^\star(\datax)) dp(\datax) + \int_{\Omega} c({T_\lambda^\star}^{-1}(\dataz), \dataz) d q(\dataz)\right] = \ot(p, q),
    \end{equation}
    where the last equality comes from the weak convergence from ${T_\lambda^\star}_\sharp p$ to $q$ and from ${{T_\lambda^\star}_\sharp}^{-1}q$ to $p$. Then combining the results of (\ref{eq:inequality1}) and (\ref{eq:inequality2}) we conclude that
    \begin{equation}
        \lim_{\lambda \rightarrow +\infty} \ot_\lambda(p, q) = \ot(p, q).
    \end{equation}
\end{proof}
Theorem \ref{thm:relationship} shows the convergence from problem $\ot_\lambda$ to $\ot$, which indicates that when the distance weight $\lambda$ goes to infinity, the relaxed problem will tend to the original OT problem. 
In the following, we show the convergence  of  the minimizer of the relaxed problem and the one of the original problem. First we define functionals $\left\{F_n(T)\right\}$ and $F(T)$ as follows:
\begin{alignat*}{4}
        F_n(T) & = \int_{\Omega}c(\datax, T(\datax)) dp(\datax) + \int_{\Omega}c(\dataz, T^{-1}(\dataz))dq(\dataz)  + n\mmd(T_{\sharp}p, q) + n\mmd(p, T^{-1}_\sharp q),\\
        F(T) & = \int_{\Omega}c(\datax, T(\datax)) dp(\datax) + \int_{\Omega}c(\dataz, T^{-1}(\dataz))dq(\dataz) + \boldone_\mathcal{E}(T) \text{, where } \mathcal{E}=\left\{T: T_\sharp p = q\right\}.
\end{alignat*}
Here, the $\boldone_\mathcal{E}$ represents the indicator function such that
\begin{equation*}
\boldone_E(T) = \left\{
\begin{aligned}
    & 0, \quad T\in \mathcal{E}, \\
    & +\infty, \quad T \notin \mathcal{E}.
\end{aligned}
\right.
\end{equation*}
Then the functional $F(T)$ can be regarded as the limit of the sequence $\{F_n(T)\}$ as $n$ goes to infinity. In the next theorem, we prove the $\Gamma$-convergence from $F_n(T)$ to $F(T)$ as $n$ goes to infinity, which reveals the relationship between the minimizers of $F_n(T)$ and the one of $F(T)$. Before giving the theorem, we first introducing a lemma about the property of the convergence of the diffeomorphism sequence, which is stated as follows:
\begin{lemma}\label{lm2}
    Suppose $\spaceT$ is a set of all the diffeomorphisms on $\Omega$ and $\{T_n\} \subset \spaceT$ is a sequence of diffeomorphisms in $\spaceT$ such that $T_n$ pointwise converges to $T$ for some mapping $T$. If $T \in \spaceT$ and the Jacobian matrix of $T_n$ is uniformly bounded by some positive constant $M$ in the sense of norm, then we can get that $\invT_n$ also pointwise converges to $\invT$. Precisely, if for all preimage  $\datax \in \Omega$, $T_n(\datax) \to T(\datax)$, then with the conditions given above, we can get that
    \begin{equation*}
        \invT_n(\dataz) \to \invT(\dataz) \text{ for all image $\dataz \in \Omega$}.
    \end{equation*}
\end{lemma}
\begin{proof}
    Suppose not, then there exists an image $z_0 \in \Omega$ such that $\invT_n(z_0) \nrightarrow \invT(z_0)$, which means that $\exists \epsilon_0 > 0$ and a sequence $\{T_n\}$ such that
    \begin{equation*}
        \|\invT_n(z_0) - \invT(z_0) \| \geq \epsilon_0.
    \end{equation*}
    Suppose for all $n$, $T_n(x_n) = z_0$ and since $\{x_n\} \subset \Omega$ is in a compact set in $\R^d$, then we can find a convergent subsequence $\{x_n\}$ (we can also denoted by $\{x_n\}$) such that $x_n \to x_0$ for some point $x_0 \in \Omega$. Then we can have that
    \begin{alignat*}{4}
        \|T_n(x_n) - T(x_0)\| & \leq \|T_n(x_n) - T_n(x_0)\| + \|T_n(x_0) - T(x_0)\|&& \\
        & = \|J_{T_n}^\top(\xi_n)(x_n - x_0)\| + \|T_n(x_0) - T(x_0)\|&& \\
        & \leq M\|x_n - x_0\| + \|T_n(x_0) - T(x_0)\| \to 0 \text{ as $n\to \infty$},&&
    \end{alignat*}
    which indicates that 
    \begin{equation*}
        z_0 = \lim_{n\to\infty} T_n(x_n) = T(x_0) \quad \Longrightarrow \quad x_0 = \invT(z_0).
    \end{equation*}
    However, by our assumption we have that
    \begin{equation*}
        \|\invT_n(z_0) - \invT(z_0) \| = \|x_n - x_0\| \geq \epsilon_0,
    \end{equation*}
    which is contradictory to $x_n \to x_0$.
\end{proof}
\begin{theorem}
    Under the space and conditions given in Lemma \ref{lm2}, $F_n(T)$ is $\Gamma$-convergent to $F(T)$. Correspondingly, suppose $T_n$ is a minimizer of $F_n(T)$ for all $n$, and any cluster point of $T_n$ in $\spaceT$ is a minimizer of $F(T)$.
\end{theorem}
\begin{proof}
    By $\Gamma$-convergence theory, we need to prove the following two results:
\begin{itemize}
    \item [i.] $\forall T_n \rightarrow T$ in $\spaceT$ pointwise, we have:
    \begin{equation*}
        F(T) \leq \liminf_{n\rightarrow +\infty} F_n(T_n).
    \end{equation*}
    \item [ii.] $\forall T \in \spaceT$, there exists $T_n \rightarrow T$ pointwise in $\spaceT$ such that
    \begin{equation*}
        F(T) \geq \limsup_{n\rightarrow+\infty} F_n(T_n).
    \end{equation*}
\end{itemize}

For ii. If $T\notin \mathcal{E}$, then we have $F(T) = +\infty$. While if $T \in \mathcal{E}$, let $T_n = T$ for $\forall n$, then we have 
\begin{alignat*}{2}
     F(T) = F_n(T_n) = \int_{\Omega} c(\datax, T(\datax))dp(\datax) + \int_\Omega c(\dataz, \invT(\dataz)) dq(\dataz).
\end{alignat*}
Therefore, for all $T\in\spaceT$ we can always choose $T_n = T$ and the inequality
\begin{equation*}
    F(T) \geq \limsup_{n\rightarrow+\infty} F_n(T_n).
\end{equation*}
always holds.

For i. If $T \notin \mathcal{E}$, then $F(T) = +\infty$ and now we'll prove one of the $n\mmd({T_{n}}_\sharp p, q)$ and $n\mmd(p, {T^{-1}_{n}}_\sharp q)$ will go to infinity so that the inequality
\begin{equation*}
    F(T) \leq \liminf_{n\rightarrow +\infty} F_n(T_n).
\end{equation*}
can hold. Here, we choose $n\mmd({T_{n}}_\sharp p, q)$ to finish our demonstration, which indicates that we'll prove
\begin{equation*}
    \liminf_{n\rightarrow+\infty} n\mmd({T_{n}}_\sharp p, q) = +\infty.
\end{equation*}
If not, there exists a subsequence $\left\{T_n\right\}$ (also denoted by $T_n$) converging to $T$  such that 
\begin{equation*}
    \lim_{n\to+\infty} \mmd({T_{n}}_\sharp p, q) = 0.
\end{equation*}
Since for all $\datax \in \Omega$, $T_n(\datax) \to T(\datax)$, then for $\forall f\in\mathcal{C}_b(\Omega)$, we have that
\begin{equation*}
\begin{aligned}
     \lim_{n\to+\infty} \mathbb{E}_{{T_n}_\sharp p} f & = \lim_{n\to+\infty} \int_\Omega f(\dataz) d {T_n}_\sharp p(\dataz) = \lim_{n\to+\infty}\int_\Omega f(T_n(\datax)) dp(\datax) \\
     & = \int_\Omega \lim_{n\to+\infty} f(T_n(\datax))dp(\datax) = \int_\Omega f(T(\datax)) dp(\datax) = \mathbb{E}_{T_\sharp p} f.
\end{aligned}
\end{equation*}
Then change of the integral and the limit derives from the Dominated Convergence Theorem since $f\in\mathcal{C}_b(\Omega)$ and $p$ is a probability measure. And this result indicates that the measure sequence $\{{T_n}_\sharp p\}$ is weak convergent to $T_\sharp p$. Then by Lemma \ref{lm1} we can conclude that
\begin{equation*}
    \lim_{n\to+\infty}\mmd({T_n}_\sharp p, T_\sharp p) = 0.
\end{equation*}
Furthermore, we obtain that
\begin{alignat*}{3}
    0\leq\mmd(T_\sharp p, q) \leq \lim_{n\to+\infty}\mmd({T_n}_\sharp p, T_\sharp p) + \lim_{n\to+\infty}\mmd({T_{n}}_\sharp p, q) = 0.
\end{alignat*}
This leads to  $\mmd(T_\sharp p, q)$ and $T_\sharp p = q$, which is contradictory to our assumption $T \notin \mathcal{E}$. Therefore we finally get that
\begin{equation*}
    \liminf_{n\rightarrow+\infty} n\mmd({T_{n}}_\sharp p, q) = +\infty.
\end{equation*}
On the other hand, if $T \in \mathcal{E}$, we have $T_\sharp p = q$. By Lemma \ref{lm2} we can first obtain that $\invT_n(\dataz)\to \invT(\dataz)$ for all $\dataz \in \Omega$. Then correspondingly, we obtain that
\begin{alignat*}{3}
     & \lim_{n\to+\infty} \int_\Omega c(\datax, T_n(\datax)) dp(\datax) + \int_\Omega c(\dataz, \invT_n(\dataz)) dq(\dataz) &\\
     & \qquad = \int_\Omega\lim_{n\to+\infty}c(\datax, T_n(\datax)) dp(\datax) + \int_\Omega\lim_{n\to+\infty}c(\dataz, \invT_n(\dataz)) dq(\dataz) &\\
     & \qquad = \int_\Omega c(\datax, T(\datax)) dp(\datax) + \int_\Omega c(\dataz, \invT_n(\dataz)) dq(\dataz).
\end{alignat*}
Similarly, the change of the integral and the limit also comes from the Dominated Convergence Theorem if we let the cost function $c(\cdot, \cdot)$ to be continuous, which is commonly used in practice. Then we can get that
\begin{alignat*}{4}
    \liminf_{n\to+\infty}F_n(T_n)
    & = \liminf_{n\to+\infty} \int_{\Omega}c(\datax, T(\datax)) dp(\datax) + \int_{\Omega}c(\dataz, T^{-1}(\dataz))dq(\dataz) && \\ 
    & \qquad\qquad\qquad\qquad + n\mmd(T_{\sharp}p, q) + n\mmd(p, T^{-1}_\sharp q) && \\
    & \geq \liminf_{n\to+\infty} \int_{\Omega}c(\datax, T(\datax))dp(\datax) + \int_{\Omega}c(\dataz, T^{-1}(\dataz))dq(\dataz) && \\ 
    & \qquad\qquad\quad + \liminf_{n\to+\infty} n\mmd(T_{\sharp}p, q) + n\mmd(p, T^{-1}_\sharp q) &&  \\
    & \geq \liminf_{n\to+\infty} \int_{\Omega}c(\datax, T(\datax))dp(\datax) + \int_{\Omega}c(\dataz, T^{-1}(\dataz))dq(\dataz) = F(T).&&
\end{alignat*}
Therefore, we obtain that $F_n(T)$ is $\Gamma$-convergence to $F(T)$. Moreover, for any $n$ let $T_n$ is a minimizer of $F_n(T)$, which means that
\begin{equation*}
    T_n = \underset{T\in\spaceT}{\arg \min} F_n(T).
\end{equation*}
Suppose $T\in\spaceT$ is a cluster point of $\{T_n\}$, then by the property of $\Gamma$-convergence we have that
\begin{equation*}
    T = \underset{T\in\spaceT}{\arg \min} F(T).
\end{equation*}
\end{proof}
Referring to \cite[Theorem~1]{kong2021universal},  the next theorem guarantees the existence of the solution to the MMD relaxation for the OT problem, which corroborates the feasibility of using MMD as the distribution distance.
\begin{theorem}\label{thm:existence}
    For any $\epsilon > 0$, there exists a $K$ and a series of invertible and continuous transformations $\left\{T_i\right\}_{i=1}^K$ such that $T = T_{K} \circ \cdots \circ T_1$ and 
    \begin{equation}
        \mmd(T_\sharp p, q) + \mmd(p, T^{-1}_\sharp q) < \epsilon.
    \end{equation}
\end{theorem}

\begin{proof}[Proof of Theorem \ref{thm:existence}]
    Define $\psi(p, q) = \mathbb{E}_{\datax\sim p}\left[\phi(\datax)\right] - \mathbb{E}_{\dataz\sim q}\left[\phi(\dataz)\right]$, then we have that
    \begin{equation}
        \mmd(p, q)^2 = \|\psi(p, q)\|^2_2.
    \end{equation}

    Suppose $f_1, f_2$ and $g_1, g_2$ are all invertible transformations, then consider the following difference:
    \begin{equation*}
        \begin{aligned}
            \Delta  & = \mmd({f_1}_\sharp p, q)^2 + \mmd(p, {g_1}_\sharp q)^2 \\
        & - \mmd(\left(f_2\circ f_1\right)_\sharp p, q)^2 - \mmd(p, \left(g_2 \circ g_1\right)_\sharp q)^2.
        \end{aligned}
    \end{equation*}
    According to Lemma \ref{mmd}, it can also be simplified as 
    \begin{equation}
        \begin{aligned}
            \Delta & = \|\E_{\datax\sim p}\left[\phi(f_1(\datax))\right] - \E_{\dataz\sim q}\left[\phi(\dataz)\right]\|_2^2 
            + \|\E_{\datax\sim p}\left[\phi(\datax)\right] - \E_{\dataz \sim q}\left[\phi(g_1(\dataz))\right]\|_2^2 \\
            & - \|\E_{\datax\sim p}\left[\phi(f_{2,1}(\datax))\right] - \E_{\dataz\sim q}\left[\phi(\dataz)\right]\|_2^2 - \|\E_{\datax\sim p}\left[\phi(\datax)\right] - \E_{\dataz\sim q}\left[\phi(g_{2,1}(\dataz))\right]\|_2^2,
        \end{aligned}
    \end{equation}
    where $f_{2,1}:=f_2\circ f_1$ and $g_{2,1}:=g_2\circ g_1$. Then the difference can be divided into two symmetrical part:
    \begin{alignat*}{3}
        && \Delta_f  = \|\E_{\datax\sim p}\left[\phi(f_1(\datax))\right] - \E_{\dataz\sim q}\left[\phi(\dataz)\right]\|_2^2 - \|\E_{\datax\sim p}\left[\phi(f_{2,1}(\datax))\right] - \E_{\dataz\sim q}\left[\phi(\dataz)\right]\|_2^2 & \\
        && \Delta_g  = \|\E_{\datax\sim p}\left[\phi(\datax)\right] - \E_{\dataz \sim q}\left[\phi(g_1(\dataz))\right]\|_2^2 - \|\E_{\datax\sim p}\left[\phi(\datax)\right] - \E_{\dataz\sim q}\left[\phi(g_{2,1}(\dataz))\right]\|_2^2&.
    \end{alignat*}
    With the symmetry, here we calculate $\Delta_f$ as an example:
    \begin{equation}
    \begin{aligned}
        \Delta_f  & =  \E_{\datax\sim p, \dataxp\sim p}\left[\phi(f_1(\datax))^\top\phi(f_1(\dataxp))\right] - 2\E_{\datax\sim p, \dataz\sim q}\left[\phi(f_1(\datax))^\top\phi(\dataz)\right]  \\
        & - \E_{\datax\sim p, \dataxp\sim p}\left[\phi(f_{2,1}(\datax))^\top \phi(f_{2,1}(\dataxp))\right] + 2\E_{\datax\sim p, \dataz\sim q}\left[\phi(f_{2,1}(\datax))^\top\phi(\dataz)\right] \\
        & = \E_{\datax\sim p, \dataxp\sim p}\left[\phi(f_1(\datax))^\top\phi(f_1(\dataxp)) - \phi(f_{2,1}(\datax))^\top \phi(f_{2,1}(\dataxp))\right] \\
        & + 2\E_{\datax\sim p, \dataz\sim q}\left[\phi(\dataz)^\top\left(\phi(f_{2,1}(\datax)) -  \phi(f_1(\datax))\right)\right]. 
    \end{aligned}
    \end{equation}
    Note that
    \begin{equation*}
        \phi(f_{2,1}(\datax)) = \phi(f_1(\datax)) + J_\phi(f_1(\datax))^\top\delta f(\datax)+ O(\|\delta f(\datax)\|^2),
    \end{equation*}
    where $\delta f(\datax) = f_{2,1}(\datax) - f_1(\datax)$. Then $\Delta_f$ can be simplified as
    \begin{equation}
        \begin{aligned}
            \Delta_f & = 2\E_{\datax\sim p, \dataz\sim q}\left[\delta f(\datax)^\top J_\phi(f_1(\datax))\phi(\dataz)\right]  \\
            & - 2\E_{\datax\sim p, \dataxp\sim p}\left[\delta f(\datax)^\top J_\phi(f_1(\datax))\phi(f_1(\dataxp))\right] + O(\|\delta f(\datax)\|^2) \\
            & = 2\E_{\datax\sim p}\left[\delta f(\datax)^\top J_\phi(f_1(\datax))\right]\cdot \left\{\E_{\dataz\sim q}\left[\phi(\dataz)\right] - \E_{\dataxp\sim p}\left[\phi(f_1(\dataxp))\right]\right\} + O(\|\delta f(\datax)\|^2) \\
            & = 2\psi(q, {f_1}_\sharp p)^\top\E_{\datax\sim p}\left[J_\phi(f_1(\datax))^\top\delta f(\datax)\right]+ O(\|\delta f(\datax)\|^2). 
        \end{aligned}
    \end{equation}
    Therefore, let $f_2$ satisfy that
    \begin{equation*}
        \delta f(\datax) = f_{2,1}(\datax) - f_1(\datax) = h J_\phi(f_1(\datax))\psi(q, {f_1}_\sharp p),
    \end{equation*}
    where the parameter $h$ can be small enough. Then we can get the lower bound of $\Delta_f$,
    \begin{equation*}
    \begin{aligned}
        \Delta_f  &\geq h\E_{\datax\sim p}\|J_\phi(f_1(\datax))\psi(q, {f_1}_\sharp p)\|_2^2 \\
         &\geq  h\E_{\datax\sim p}\|J_\phi(f_1(\datax_0))\psi(q, {f_1}_\sharp p)\|_2^2 \\
         &\geq h\sigma \|\psi(q, {f_1}_\sharp p)\|_2^2,
    \end{aligned}
    \end{equation*}
    where $\datax_0$ represents
    \begin{equation*}
        \datax_0 = \arg\min_{\datax}\E_{\datax\sim p}\|J_\phi(f_1(\datax))\psi(q, {f_1}_\sharp p)\|_2^2,
    \end{equation*}
    and correspondingly, $\sigma$ is the smallest eigenvalue of the matrix $J_\phi(\datay)^\top J_\phi(\datay)$ for any $\datay \in \R^d$. In practice we can choose the kernel elaborately to make sure that $\sigma > 0$ (for example, Gaussian kernel). Then we get that
    \begin{equation}
        \Delta_f \geq  h\sigma\|\psi(q, {f_1}_\sharp p)\|_2^2  = h\sigma\mmd({f_1}_\sharp p, q)^2
    \end{equation}
    Therefore, if we define
    \begin{equation*}
        S_k = \mmd((f_k\circ \cdots \circ f_0)_\sharp p, q)^2,
    \end{equation*}
    where $f_0$ represents the identity mapping $I_d$. Then we get that for any $k > 0$,
    \begin{equation}
        S_{k-1} - S_{k} \geq h_k\sigma S_{k-1} \Leftarrow S_k \leq (1 - h_k\sigma)S_{k-1}.
    \end{equation}
    Hence, for any $\epsilon > 0$, let each $h_k < C_0$ for some positive constant $C_0$ and $K > O(\log(\epsilon)/ \log(1 - C_0\sigma))$ , then we have
    \begin{alignat}{3}
    S_K \leq (1 - C_0\sigma)S_{K-1} \leq \cdots \leq (1 - C_0\sigma)^K S_0 = (1 - C_0\sigma)^K \mmd(p, q)^2 < \epsilon.  
    \end{alignat}
    Therefore, if we choose $T_i = f_i$ for $i=1,\cdots,K$ and define $T = T_K \circ \cdots \circ T_1$, then it follows that
    \begin{equation}
        \mmd(T_\sharp p, q) < \epsilon.
    \end{equation}
    On the other hand, since $T$ is a diffeomorphism,
    by the definition of $\text{MMD}(p, T^{-1}_\sharp q)$ we can obtain that
    \begin{equation}
        \begin{aligned}
            \text{MMD}(p, T^{-1}_\sharp q) & = \sup_{\|f\|_{\spaceH}\leq 1}\mathbb{E}_p[f(\mathbf{x})] - \mathbb{E}_{T^{-1}_\sharp q}[f(\mathbf{y})] \\
            & = \sup_{\|f\|_{\spaceH}\leq 1}\mathbb{E}_{T_\sharp p}[f(T^{-1}(\mathbf{x}))] - \mathbb{E}_{q}[f(T^{-1}(\mathbf{y}))] \\
            & \leq \sup_{\|f\|_{\spaceH}\leq 1}\mathbb{E}_{T_\sharp p}[f(\mathbf{x})] - \mathbb{E}_{q}[f(\mathbf{y})] \\
            & = \text{MMD}(T_\sharp p, q) < \epsilon.
        \end{aligned}
    \end{equation}
    The inequality is because $f\circ \invT$ still belongs to $\spaceH$ since $T\invT$ is also a diffeomorphism. Therefore, with simple scaling we finally get that
    \begin{equation}
        \mmd(T_\sharp p, q) + \mmd(p, T^{-1}_\sharp q) < \epsilon.
    \end{equation}
\end{proof}
With Theorem \ref{thm:existence}, it is feasible to obtain an optimal transformation between $p$ and $q$ through a series of invertible normalizing flow modules such as RealNVP\cite{dinh2016density}, Glow\cite{kingma2018glow}, etc.

\section{Experiments on Toy and Image Datasets}\label{experiments}
\subsection{Implementation details} For all the experiments, we use FrEIA \cite{freia} to build invertible flow structure. Moreover, in the calculation of $\mmd_b$, we use multi-kernel MMD,i.e. a  weighted combination of multiple kernels \cite[Section~2.4]{gretton2012optimal}. For all the experiments, the flow contains 8 
INN blocks with fully connected or convolutional subnets. The batch size is equal to 200 and the optimizer is chosen as AdamW \cite{loshchilov2017decoupled}.

For each pair of two datasets, the weight parameter $\beta$ in the empirical loss \eqref{loss:empirical} is a hyperparameter which has been fine-tuned for the best performance. Besides, we also have some ablation experiments about the influence of weight $\beta$ and the symmetrical design to the performance of the optimal solution learned by INN.

\subsection{Illustrative examples} To validate the effectiveness of the  proposed approach, we simulate four pairs of illustrative examples in $\R^2$. The proposed SyMOT-Flow method is compared with the model only with MMD (single MMD with $\beta=0$ in \eqref{loss:empirical}) and SWOT-Flow proposed in \cite{coeurdoux2023learning}.

\begin{figure}[htbp]
	\begin{center}
	\scalebox{.85}[.85]{
		\begin{tabular}{c@{\vspace{-6pt}}c@{\vspace{2pt}}c@{\vspace{-8pt}}c@{\vspace{-10pt}}c@{\hspace{1pt}}c@{\hspace{1pt}}c}
		   & Training Data & \multicolumn{1}{c}{SWOT-Flow}& \multicolumn{1}{c}{ MMD}&   \multicolumn{1}{c}{SyMOT-Flow}\\
			\put(-10,0){\rotatebox{90}{\textbf{Example 1}} }  &
			\includegraphics[width=.25\linewidth,height=.25\linewidth]{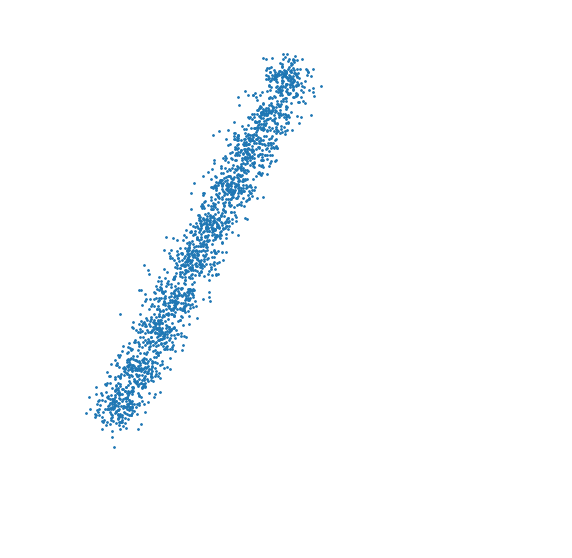}&	
			\includegraphics[width=.25\linewidth,height=.25\linewidth]{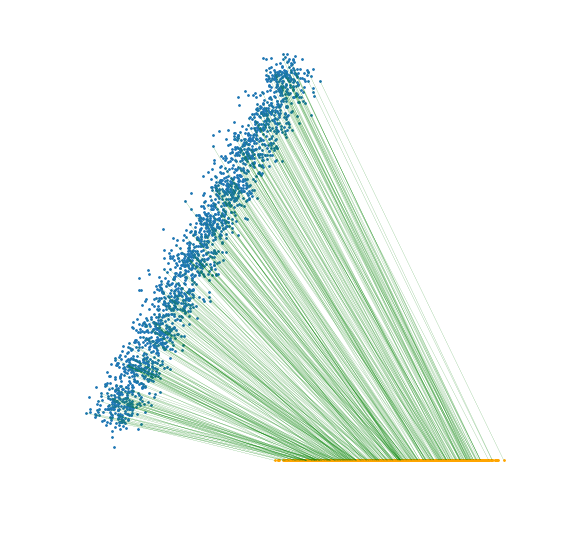}&
            \includegraphics[width=.25\linewidth,height=.25\linewidth]{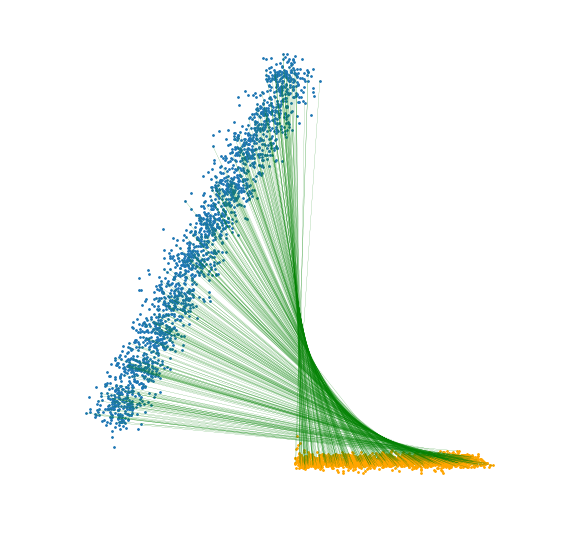}&
            \includegraphics[width=.25\linewidth,height=.25\linewidth]{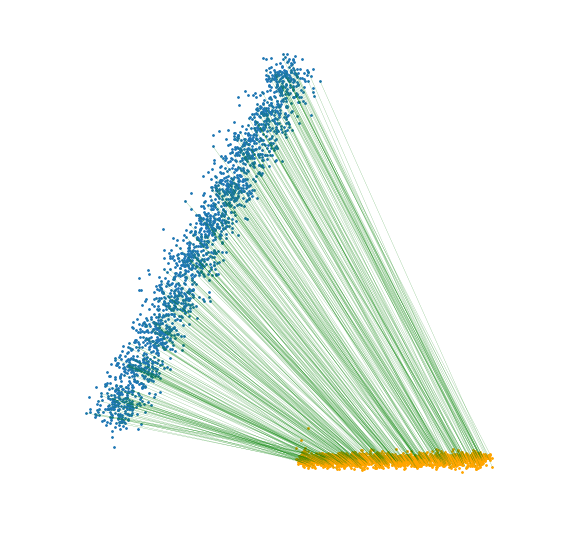}\\
          & \includegraphics[width=.25\linewidth,height=.25\linewidth]{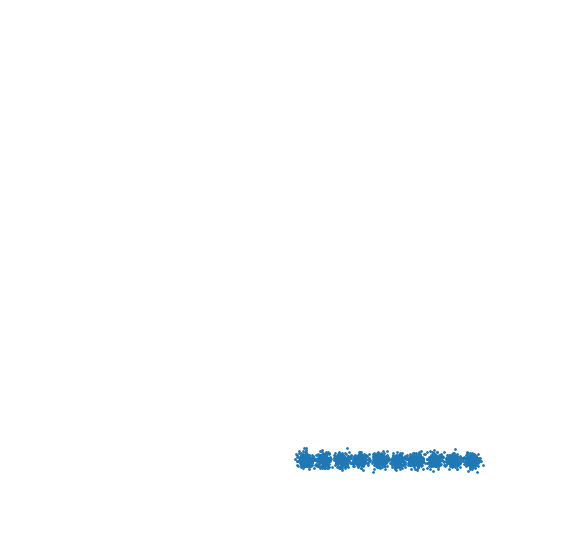}&	
             \includegraphics[width=.25\linewidth,height=.25\linewidth]{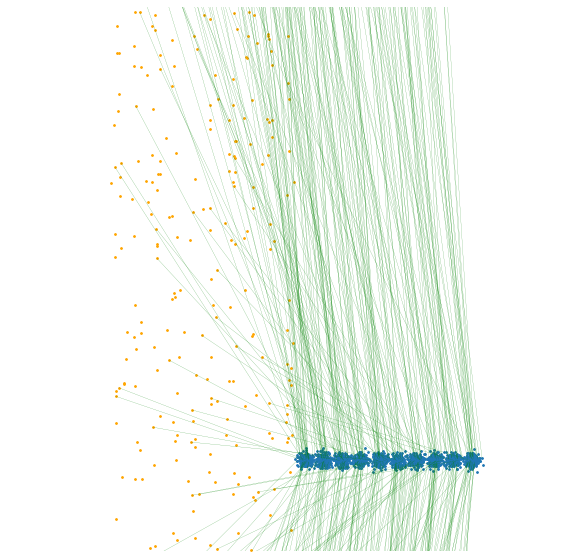}&
            \includegraphics[width=.25\linewidth,height=.25\linewidth]{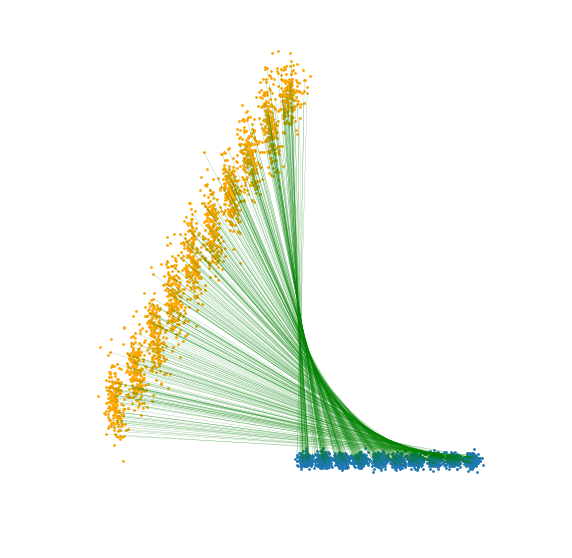}&
            \includegraphics[width=.25\linewidth,height=.25\linewidth]{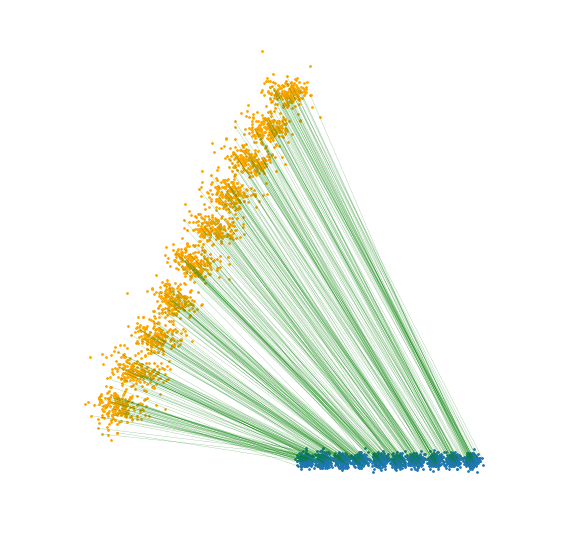}\\
        \put(-10,0){\rotatebox{90}{\textbf{Example 2}} } &
		\includegraphics[width=.25\linewidth,height=.25\linewidth]{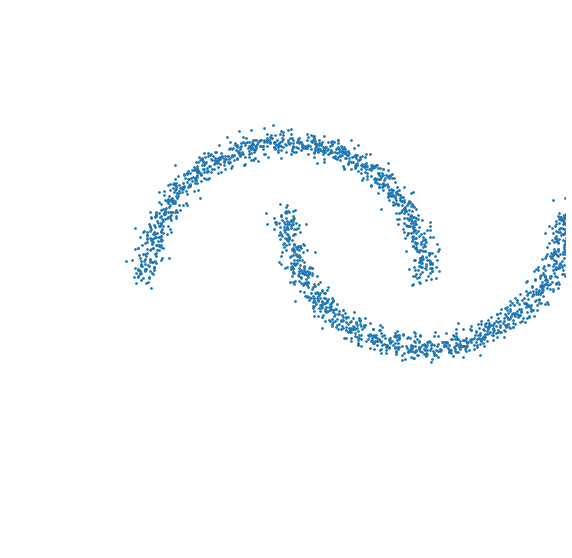}&
			\includegraphics[width=.25\linewidth,height=.25\linewidth]{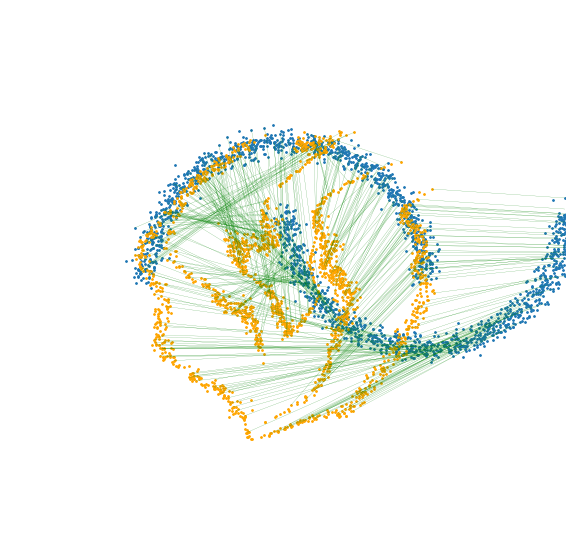}&
			\includegraphics[width=.25\linewidth,height=.25\linewidth]{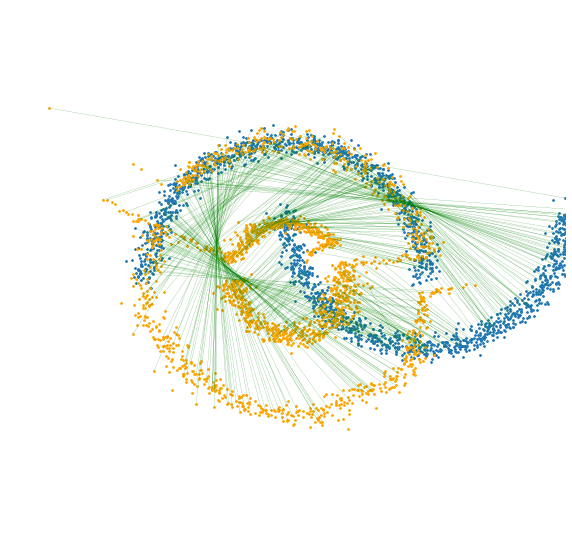}&
			\includegraphics[width=.25\linewidth,height=.25\linewidth]{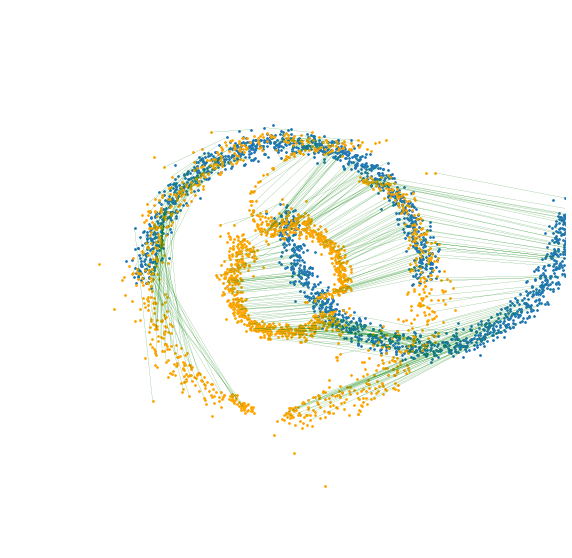}\\
		&	\includegraphics[width=.25\linewidth,height=.25\linewidth]{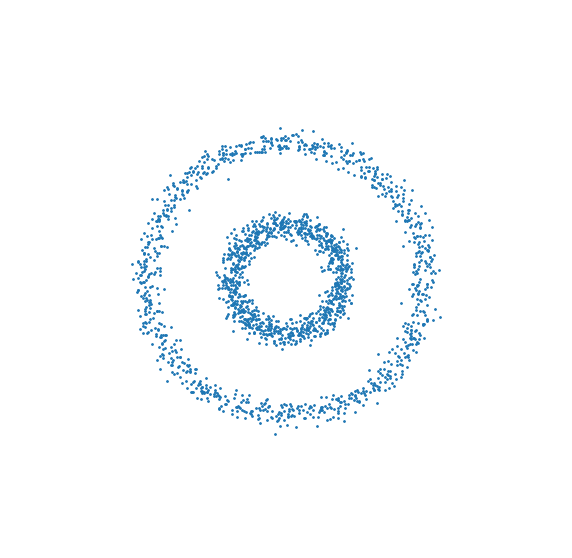}&
			\includegraphics[width=.25\linewidth,height=.25\linewidth]{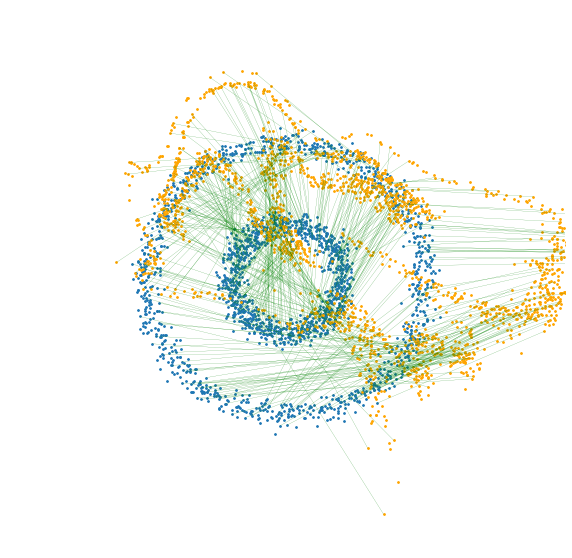}&
			\includegraphics[width=.25\linewidth,height=.25\linewidth]{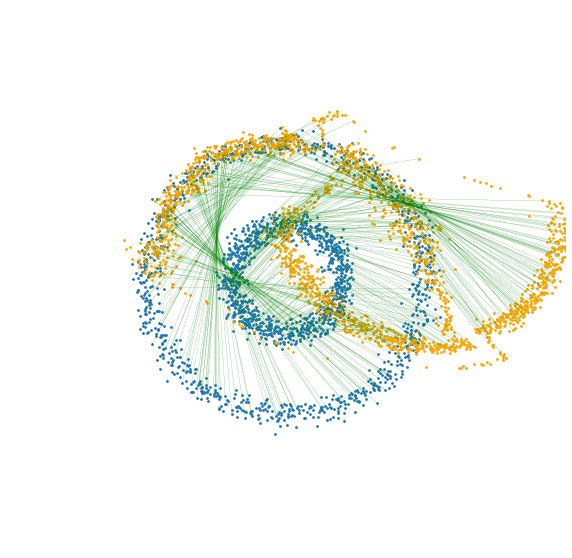}&
			\includegraphics[width=.25\linewidth,height=.25\linewidth]{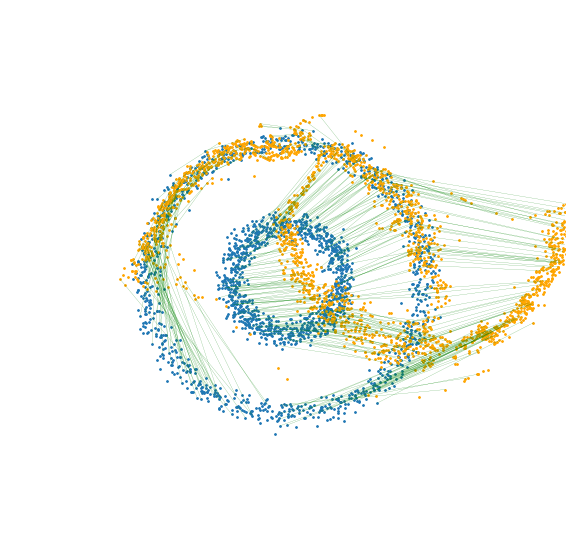}
		\end{tabular}
		}
		\caption { The input (blue) and generated points (orange) by learned map (green lines) between two  sets of training data with different methods.  }
		\label{fig:VisualResult}
	\end{center}
\end{figure}

Fig. \ref{fig:VisualResult} shows two sets of experiments. In all the experiments, we randomly sample $2000$ points $\{\datax_i\}$ and $\{\dataz_j\}$ from each distribution, represented by  blue points in the two rows respectively. For the first one, the two sets of points are drawn from a series of Gaussian distributions with the centers along two lines with different covariance respectively. And for the second one, the data are sampled from two-moon and circles, with noise variance $0.05$.

In each sub-figure of Fig. \ref{fig:VisualResult}, the blue points represent the original samples and the orange ones represent the generated data $\left\{T_\theta(\datax_i)\right\}$ and $\left\{{T_\theta}^{-1}(\dataz_j)\right\}$ via the learned mappings $T_\theta$ and $T_\theta^{-1}$ (the correspondence are linked with green lines) by different methods. It can be seen that with  the OT regularization, the method tends to learn a map which gives the shortest distance from one sets to the another, comparing to the single MMD and SWOT-Flow methods. Moreover, the symmetric loss provides a more stable sampling for both forward and inverse transformation. More visual images/MRI on different test sets are given in supplementary material.

Numerically, the corresponding OT cost and MMD distances are shown in Tab. \ref{tab:ot_cost} for each method and each pair of data. SyMOT-Flow obtains  smaller OT cost and MMD distance for forward and backward of flow in comparison with single MMD and SWOT-Flow, as expected.
\begin{table*}[h!]
    \caption{The OT cost and MMD distance of forward/backward direction of flow.}
    \label{tab:ot_cost}
    \centering
    \renewcommand{\arraystretch}{1.2}
    \begin{tabular}{lllll}
    \hline
      & Method  & SWOT-Flow \cite{coeurdoux2023learning} & single MMD & SyMOT-Flow \\
      \hline
    \multirow{4}{*}{\makecell[l]{OT\\ Cost}}&Moons to Circles & 0.684/0.723    &0.924/0.889 &\textbf{0.295}/\textbf{0.288} \\
          & Gauss to Gauss & 32.924/32.940   &35.310/33.956   & \textbf{32.459}/\textbf{32.475} \\
         & 8 Gauss to 8 Gauss &7.903/7.924 & 17.427/17.799& \textbf{7.539}/\textbf{7.483} \\
          &  Linear Gauss &  \textbf{138.580}/5834.745   & 156.393/153.093&139.780/\textbf{134.318} \\
          \hline
    \multirow{4}{*}{\makecell[l]{MMD\\ distance}} &    Moons to Circles &1.7e-2/4.3e-2     & 6.3e-3/5.7e-3& \textbf{2.9e-3}/\textbf{2.7e-3} \\
        &Gauss to Gauss &  6.6e-2/6.6e-2   & \textbf{2.3e-2}/\textbf{1.8e-2}& 6.6e-2/6.3e-2 \\
        &8 Gauss to 8 Gauss & 1.4e-2/1.4e-2 &  7.9e-3/3.9e-3& \textbf{4.2e-3}/\textbf{2.5e-3} \\
        &Linear Gauss &  5.6e-3/3.5   & 3.3e-3/7.0e-3& \textbf{1.1e-3}/\textbf{1.3e-3} \\
    \hline
    \end{tabular}
\end{table*}
\subsection{Ablation Study} 
\begin{figure}[htbp]
	\centering
	\scalebox{.9}[.9]{
		\begin{tabular}{c@{\vspace{0pt}}c@{\vspace{-7pt}}c@{\vspace{3pt}}c@{\vspace{-10pt}}c@{\hspace{1pt}}c@{\hspace{1pt}}c}
		    Training Data & \multicolumn{1}{c}{One Direction}&   \multicolumn{1}{c}{SyMOT-Flow}&\\
			\includegraphics[width=.33\linewidth,height=.33\linewidth]{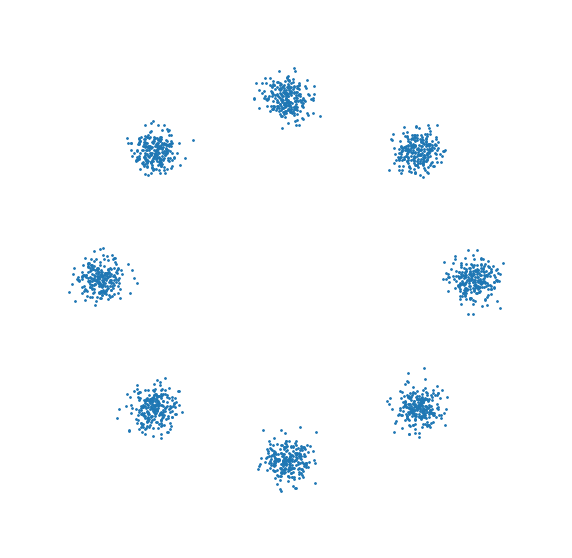}&	
			\includegraphics[width=.33\linewidth,height=.33\linewidth]{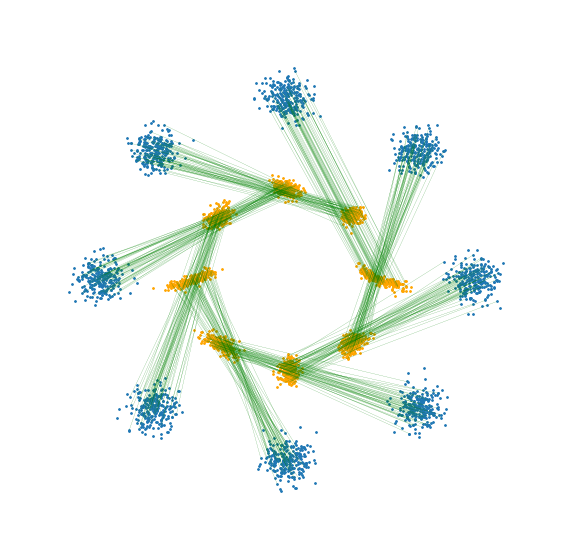}&
            \includegraphics[width=.33\linewidth,height=.33\linewidth]{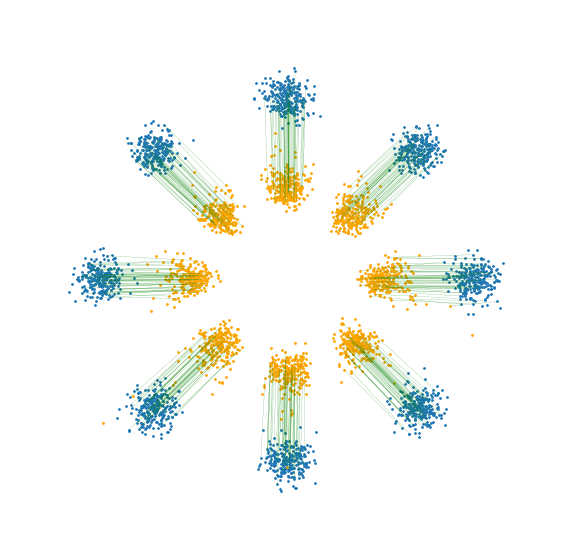}\\
            \includegraphics[width=.33\linewidth,height=.33\linewidth]{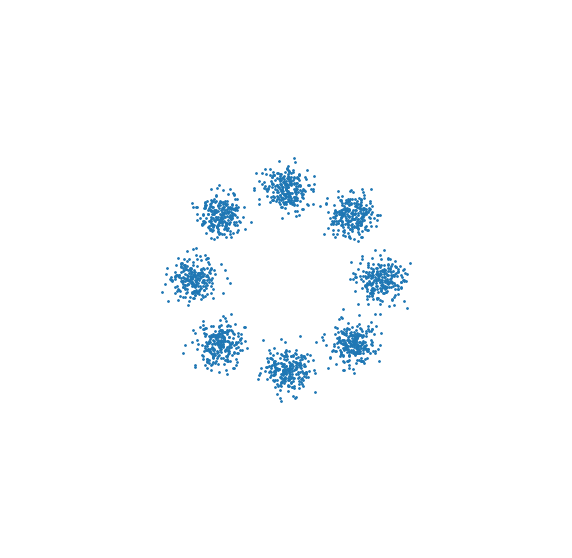}&
            \includegraphics[width=.33\linewidth,height=.33\linewidth]{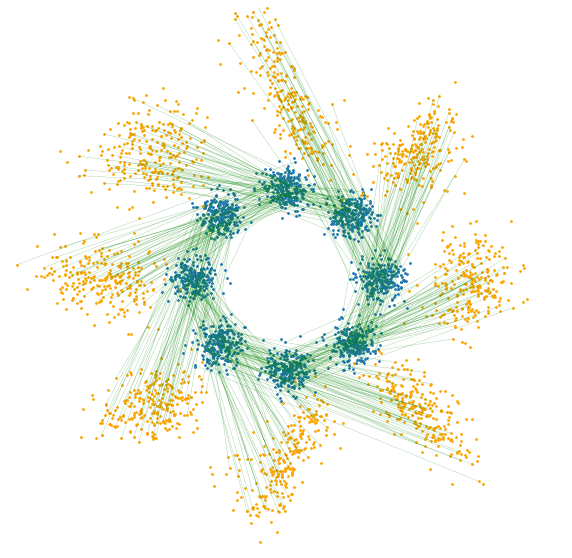}&	
            \includegraphics[width=.33\linewidth,height=.33\linewidth]{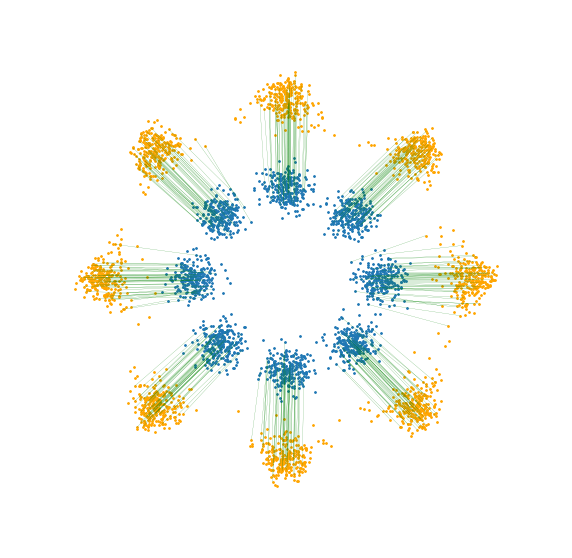}
		\end{tabular}
		}
		\caption { The input $\datax$ (blue) and generated points $\dataz$ (orange) by learned map $T_\theta$ (green lines) between two  sets of training data with and without the reversed flow loss.  }
	\label{VisualResult}
\end{figure}

To assess the impact of the symmetric reverse flow component, experiments are conducted and the results are shown in Fig. \ref{VisualResult} for the case with and without the reversed flow loss. It can be seen that in the absence of the reversed component, the generated samples exhibits a higher level dissimilarity to the intrinsic distribution, compared to  those generated with the proposed symmetrical loss.

\begin{figure}[htbp]
    \centering
    \begin{minipage}{0.35\textwidth}
        \includegraphics[width=1.0\textwidth]{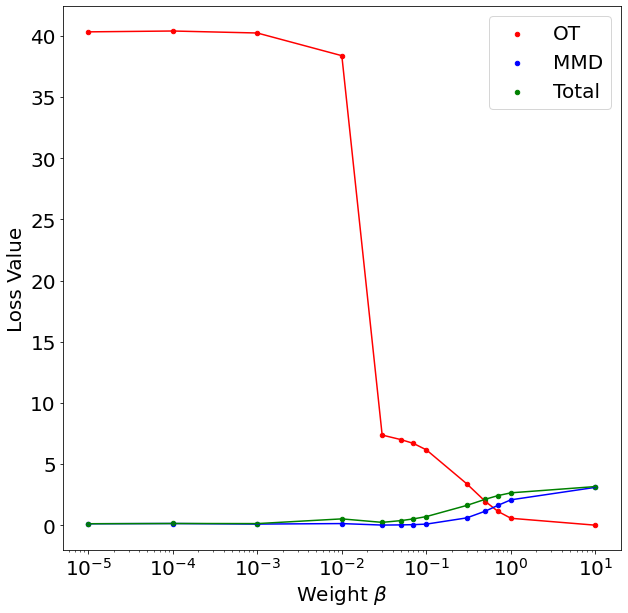}
    \end{minipage}
    \hspace{1em}
    \begin{minipage}{0.6\textwidth}
    \scalebox{0.8}[0.8]{
        \begin{tabular}{c@{\hspace{2pt}}|c@{\hspace{3pt}}c@{\hspace{3pt}}c@{\hspace{3pt}}c@{\hspace{3pt}}c@{\hspace{3pt}}c@{\hspace{3pt}}c@{\hspace{3pt}}c@{\hspace{2pt}}c@{\hspace{2pt}}c}
			\hline
			Weight $\beta$& 1e-5   & 1e-4  & 1e-3   & 1e-2  & 1e-1  & 1e0  & 1e1\\
			OT            & 40.321 &40.390 & 40.228 &38.501 & 6.134 & 0.577 &\textbf{0.006} \\
			MMD           & 0.124  &0.142  &0.097   & 0.256 &\textbf{0.096}&2.071&3.099\\
			\hline
			Weight $\beta$& 3e-2   & 5e-2  & 7e-2   & 1e-1  & 3e-1  & 5e-1  & 7e-1\\
			OT            & 7.371  &7.000  & 6.694  &6.163  & 3.392 & 1.940 & \textbf{1.142} \\
			MMD           & \textbf{0.016} &0.031   &0.050  & 0.095 & 0.603 &1.158&1.622\\
			\hline	
		\end{tabular}}
    \end{minipage}
    \caption{The value of MMD distance and OT cost with increasing weight parameter $\beta$ from $10^{-5}$ to $10$.}
    \label{fig:weight}
\end{figure}

The selection of the weight parameter $\beta$ for the OT regularization plays a crucial role in achieving superior performance in learning the optimal transport. The change of values of MMD and OT referring to several different $\beta$ is shown in the left figure in \ref{fig:weight} and the corresponding OT costs and MMD distances are presented in the right table in \ref{fig:weight}. The red line corresponds to the values of the OT cost, the blue line represents the results of the MMD distance, and the green line indicates the total loss. The weight parameter $\beta$ varies from $10^{-5}$ to $10$ with 10 logarithmically spaced increments, more refined results are displayed. It is obvious as $\beta$ increases, the  OT cost decreases. Meanwhile, it is crucial to maintain a close-to-zero MMD throughout this process, as an excessively large $\beta$ would result in the learned transformation being  an identity map. More visualized results are provided in supplementary.

\subsection{MINST and Fashion-MINST}
\begin{figure*}[htbp]
	\begin{center}
	\scalebox{0.95}[0.95]{
		\begin{tabular}{c@{\hspace{2pt}}c@{\hspace{2pt}}c@{\hspace{2pt}}c@{\hspace{2pt}}c@{\hspace{2pt}}c@{\hspace{2pt}}c@{\hspace{2pt}}c@{\hspace{2pt}}c@{\hspace{2pt}}c}
		& \textbf{Input} & \textbf{Generated}\\
		\put(-10,40){ \rotatebox{90}{\textbf{MNIST}} }&
		  \includegraphics[width=.45\textwidth]{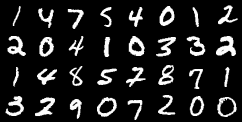}&
		  \includegraphics[width=.45\textwidth]{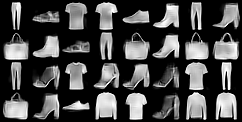}\\
		\put(-10,20){ \rotatebox{90}{\textbf{Fashion-MNIST}} }&
		   \includegraphics[width=.45\textwidth]{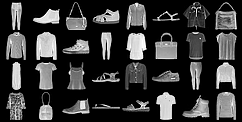}&
		   \includegraphics[width=.45\textwidth]{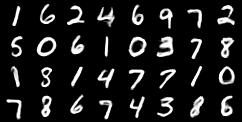}
		\end{tabular}}
		    \caption{The generation results between MNIST and Fashion-MNIST datasets}
		\label{Mnist}
	\end{center}
\end{figure*}

We evaluate SyMOT-Flow on  two sets of high dimension data: MNIST and Fashion-MNIST. 
Both of MNIST and FMNIST contain $60000$ training data and $10000$ test data. In the training of MNIST-to-FMNIST, we use the encoder pretrained on ImageNet and pretrained  a two-layer upsample-conv decoder, which is fixed in the training of flow structures and SyMOT-Flow is applied to learn the transformation in the feature spaces. The size of feature shape is $(256, 7, 7)$ and the flow is trained for 200 epochs via optimizer AdamW with learning rate $10^{-4}$ and weight decay $10^{-5}$. The weight of the OT penalty is $10^{-4}$ and the figures show the transferred results on the test dataset. Fig. \ref{Mnist} show the results of the generated samples between  MNIST and Fashion-MNIST datasets with the learned transformation of SyMOT-Flow. It can be seen that SyMOT-Flow can generate high quality data for both MNISt and Fashion-MNIST datasets.
 
\subsection{Medical bi-modal images transformation}
Except the performance on the MNIST-class datasets, we also apply our SyMOT-Flow to the transformation between medical image modalities:  MRI T1/T2 and CT/MRI, which has higher dimension than the MNIST-class datasets. We make the comparison to other SOTA style transfer methods such as CycleGAN \cite{zhu2017unpaired}, ArtFlow \cite{an2021artflow}, UNIT \cite{liu2017unsupervised} based on the SSIM \cite{wang2004image} and FID scores \cite{heusel2017gans}. 

The datasets for MRI T1/T2 transformation are slices extracted from the BraTS 3D MRI images \cite{baid2021rsna, menze2014multimodal}. Both T1 and T2 datasets contain \textbf{1000} images for training and \textbf{251} for testing. In our model, we use vq-vae-2\cite{razavi2019generating} to encode both of the modalites to the corresponding feature spaces respectively. The size of original MRI images is $(192, 192)$ and the sizes of the feature spaces after vq-vqe-2 are $(128, 24, 24)$ and $(128, 48, 48)$. We trained our flow model for 200 epochs via optimizer AdamW with initial learning rate $10^{-3}$ and weight decay $10^{-4}$. The layer of flow is 4 and the structure we used comes from Glow, in which the subnet is designed as a eight-block residual net. The weight of the OT penalty is $10^{-5}$. For all the models, the batch size is 32 and the time cost refers to the whole training time. Especially, for ArtFlow it contains both the time training from T1-to-T2 and T2-to-T1, since it does not have reversibility and can only achieve transfer from one style to another at a time. And for our model, it also contains the time of pretraining the decoder. As we can see, our model performs best on T2 and has the best SSIM on both T1 and T2, with almost the least training time. On T1, we note that although ArtFlow has the lowest FID, the transferred results have intensity mismatch on some regions comparing to the ground truth, which is not natural for T1 image. Moreover, although the performances between UNIT and our model are closed on T1, our model still outperforms UNIT on T2 significantly. The results show that our methods outperformed other GAN or flow based method both qualitatively and  quantitatively which can also scale well on high dimensional data in terms of computation time. 

\begin{figure*}[!htp]
	\begin{center}
        \scalebox{0.95}[0.95]{
	\setlength\tabcolsep{1pt}
		\begin{tabular}{c@{\hspace{2pt}}c@{\hspace{2pt}}c@{\hspace{2pt}}c@{\hspace{2pt}}c@{\hspace{2pt}}:c@{\hspace{2pt}}c@{\hspace{2pt}}c@{\hspace{2pt}}c@{\hspace{2pt}}}
		& \multicolumn{4}{c}{\textbf{MRI-T1}} & \multicolumn{4}{c}{\textbf{MRI-T2}}\\
		  \put(-8,15){\rotatebox{90}{\textbf{\small{GT}}}}&
		  \includegraphics[width=.12\textwidth]{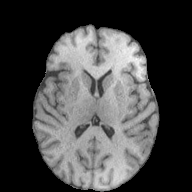}&
		  \includegraphics[width=.12\textwidth]{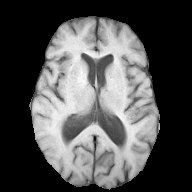}&
		  \includegraphics[width=.12\textwidth]{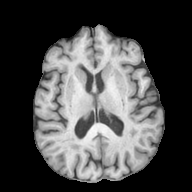}&
		  \includegraphics[width=.12\textwidth]{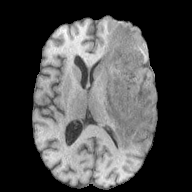}&
		  \includegraphics[width=.12\textwidth]{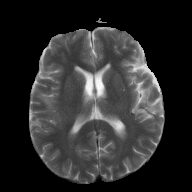}&
		  \includegraphics[width=.12\textwidth]{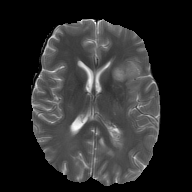}&
		  \includegraphics[width=.12\textwidth]{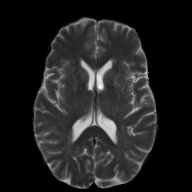}&
		  \includegraphics[width=.12\textwidth]{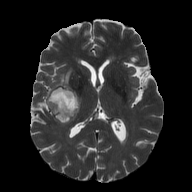}\\
		  \hdashline
		  \specialrule{0em}{2pt}{2pt}\put(-8,0){\rotatebox{90}{\textbf{\small{CycleGAN}}}}&
		  \includegraphics[width=.12\textwidth]{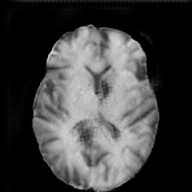}&
		  \includegraphics[width=.12\textwidth]{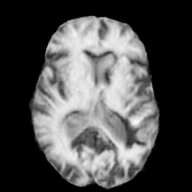}&
		  \includegraphics[width=.12\textwidth]{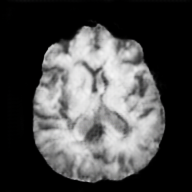}&
		  \includegraphics[width=.12\textwidth]{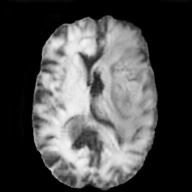}&
		  \includegraphics[width=.12\textwidth]{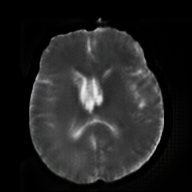}&
		  \includegraphics[width=.12\textwidth]{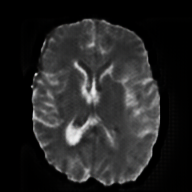}&
		  \includegraphics[width=.12\textwidth]{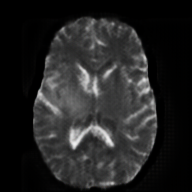}&
		  \includegraphics[width=.12\textwidth]{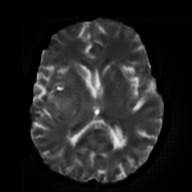}\\
		  \put(-8,8){\rotatebox{90}{\textbf{\small{ArtFlow}}}}&
	      \includegraphics[width=.12\textwidth]{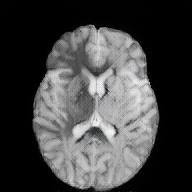}&
		  \includegraphics[width=.12\textwidth]{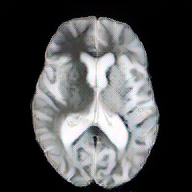}&
		  \includegraphics[width=.12\textwidth]{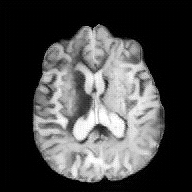}&
		  \includegraphics[width=.12\textwidth]{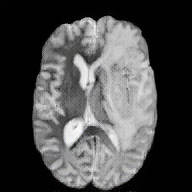}&
		  \includegraphics[width=.12\textwidth]{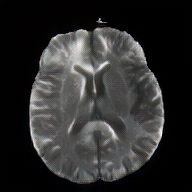}&
		  \includegraphics[width=.12\textwidth]{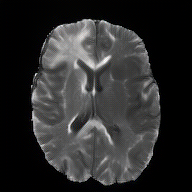}&
		  \includegraphics[width=.12\textwidth]{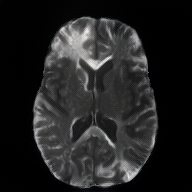}&
		  \includegraphics[width=.12\textwidth]{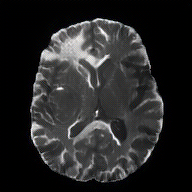}\\
		  \put(-8,10){\rotatebox{90}{\textbf{\small{UNIT}}}}& 
		  \includegraphics[width=.12\textwidth]{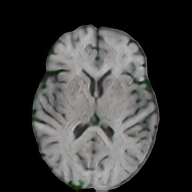}&
		  \includegraphics[width=.12\textwidth]{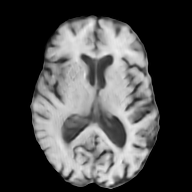}&
		  \includegraphics[width=.12\textwidth]{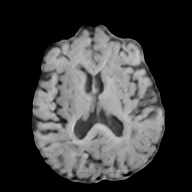}&
		  \includegraphics[width=.12\textwidth]{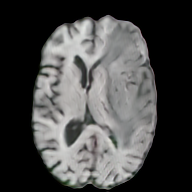}&
		  \includegraphics[width=.12\textwidth]{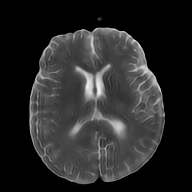}&
		  \includegraphics[width=.12\textwidth]{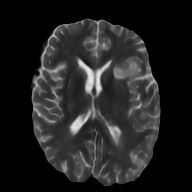}&
		  \includegraphics[width=.12\textwidth]{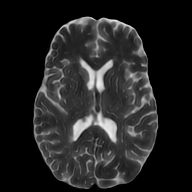}&
		  \includegraphics[width=.12\textwidth]{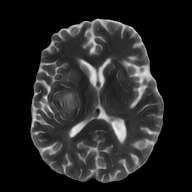}\\
		  \put(-8,15){\rotatebox{90}{\textbf{\small{Ours}}}}& 
		  \includegraphics[width=.12\textwidth]{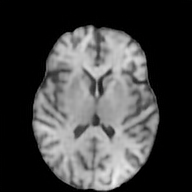}&
		  \includegraphics[width=.12\textwidth]{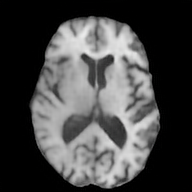}&
		  \includegraphics[width=.12\textwidth]{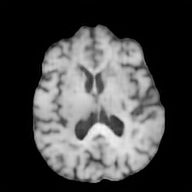}&
		  \includegraphics[width=.12\textwidth]{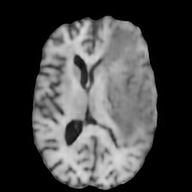}&
		  \includegraphics[width=.12\textwidth]{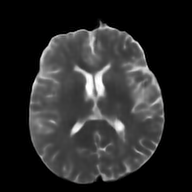}&
		  \includegraphics[width=.12\textwidth]{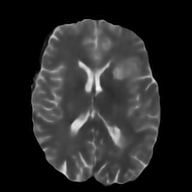}&
		  \includegraphics[width=.12\textwidth]{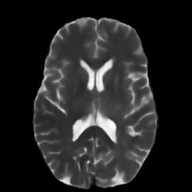}&
		  \includegraphics[width=.12\textwidth]{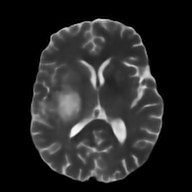}\\
		\end{tabular}}
	\end{center}
	\caption{The transferred results between MRI T1 and T2 for different methods.}
	\label{fig:mri-t1-t2}
\end{figure*}

\begin{figure*}[h!]
	\begin{center}
        \scalebox{0.95}[.95]{
	\setlength\tabcolsep{1pt}
		\begin{tabular}{c@{\hspace{2pt}}c@{\hspace{2pt}}c@{\hspace{2pt}}c@{\hspace{2pt}}c@{\hspace{2pt}}:c@{\hspace{2pt}}c@{\hspace{2pt}}c@{\hspace{2pt}}c@{\hspace{2pt}}}
		& \multicolumn{4}{c}{\textbf{CT}} & \multicolumn{4}{c}{\textbf{MR}}\\
		  \put(-8,30){\rotatebox{90}{\textbf{GT}}}&
		  \includegraphics[width=.12\textwidth]{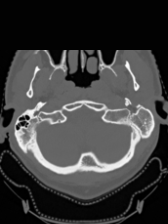}&
		  \includegraphics[width=.12\textwidth]{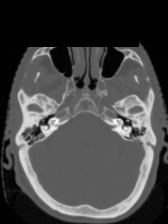}&
		  \includegraphics[width=.12\textwidth]{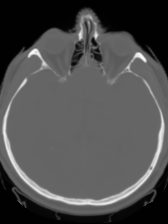}&
		  \includegraphics[width=.12\textwidth]{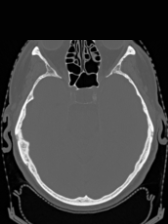}&
		  \includegraphics[width=.12\textwidth]{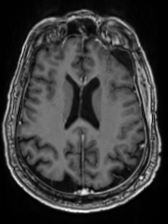}&
		  \includegraphics[width=.12\textwidth]{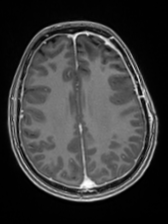}&
		  \includegraphics[width=.12\textwidth]{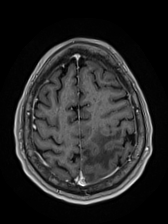}&
		  \includegraphics[width=.12\textwidth]{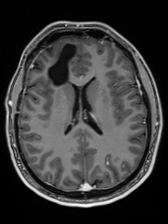}\\
		  \hdashline
		  \specialrule{0em}{2pt}{2pt}\put(-8,15){\rotatebox{90}{\textbf{CycleGAN}}}&
        \includegraphics[width=.12\textwidth]{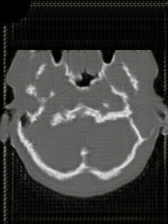}&
        \includegraphics[width=.12\textwidth]{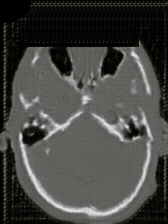}&
        \includegraphics[width=.12\textwidth]{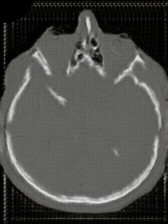}&
        \includegraphics[width=.12\textwidth]{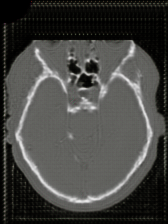}&
        \includegraphics[width=.12\textwidth]{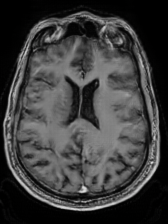}&
        \includegraphics[width=.12\textwidth]{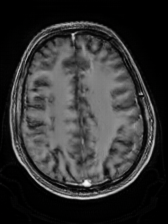}&
        \includegraphics[width=.12\textwidth]{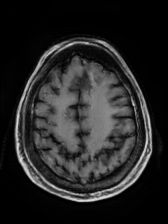}&
        \includegraphics[width=.12\textwidth]{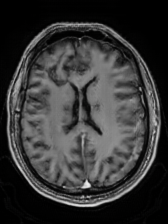}\\
        \put(-8,18){\rotatebox{90}{\textbf{ArtFlow}}}&
        \includegraphics[width=.12\textwidth]{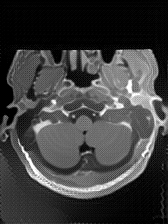}&
        \includegraphics[width=.12\textwidth]{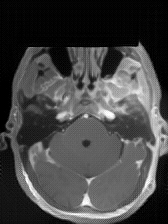}&
        \includegraphics[width=.12\textwidth]{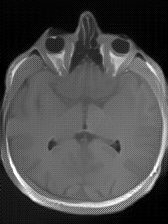}&
        \includegraphics[width=.12\textwidth]{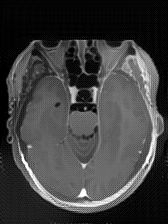}&
        \includegraphics[width=.12\textwidth]{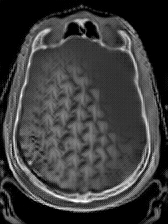}&
        \includegraphics[width=.12\textwidth]{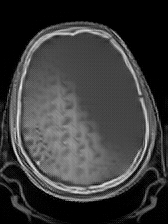}&
        \includegraphics[width=.12\textwidth]{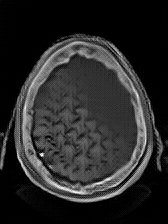}&
        \includegraphics[width=.12\textwidth]{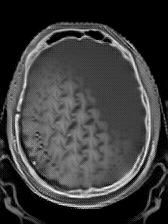}\\
        \put(-8,25){\rotatebox{90}{\textbf{UNIT}}}& 
        \includegraphics[width=.12\textwidth]{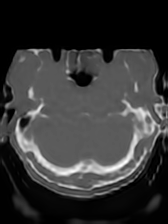}&
        \includegraphics[width=.12\textwidth]{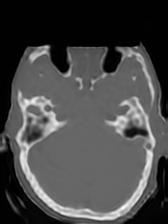}&
        \includegraphics[width=.12\textwidth]{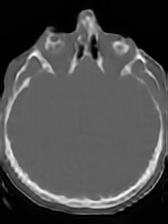}&
        \includegraphics[width=.12\textwidth]{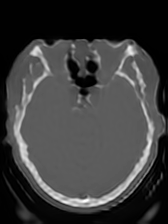}&
        \includegraphics[width=.12\textwidth]{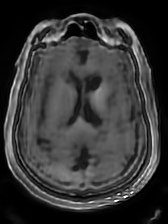}&
        \includegraphics[width=.12\textwidth]{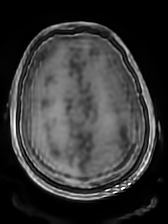}&
        \includegraphics[width=.12\textwidth]{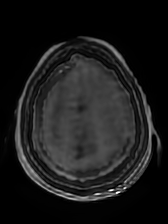}&
        \includegraphics[width=.12\textwidth]{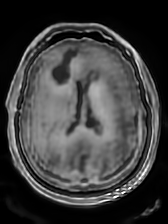}\\
        \put(-8,25){\rotatebox{90}{\textbf{Ours}}}& 
        \includegraphics[width=.12\textwidth]{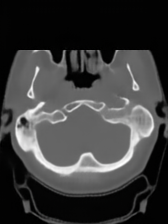}&
        \includegraphics[width=.12\textwidth]{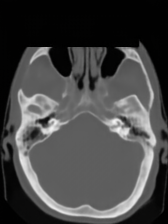}&
        \includegraphics[width=.12\textwidth]{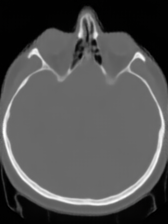}&
        \includegraphics[width=.12\textwidth]{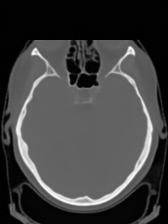}&
        \includegraphics[width=.12\textwidth]{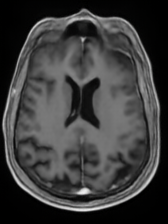}&
        \includegraphics[width=.12\textwidth]{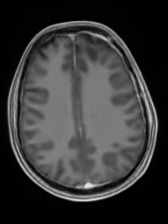}&
        \includegraphics[width=.12\textwidth]{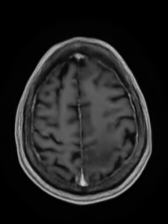}&
        \includegraphics[width=.12\textwidth]{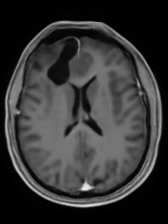}\\
		\end{tabular}}
	\end{center}
	\caption{The transferred results between CT and MR for different methods.}
	\label{fig:ct-mr}
\end{figure*}
\begin{table*}[h!]
    \centering
    \caption{The comparisons between our methods with other SOTA style transfer methods}
    \scalebox{.95}[.95]{
    \renewcommand{\arraystretch}{1.2}
    \begin{tabular}{|c|c|c|c|c|c|c|c|c|}
    \hline
        \multirow{3}{*}{Model} & \multicolumn{4}{c|}{MRI T1 \& T2} & \multicolumn{4}{c|}{CT \& MRI} \\
    \cline{2-9}
        & \multicolumn{2}{c|}{T1} & \multicolumn{2}{c|}{T2} & \multicolumn{2}{c|}{CT} & \multicolumn{2}{c|}{MRI} \\
    \cline{2-9}
        & SSIM $\uparrow$ & FID $\downarrow$ & SSIM $\uparrow$ & FID $\downarrow$ & SSIM $\uparrow$ & FID $\downarrow$ & SSIM $\uparrow$ & FID $\downarrow$ \\
    \hline
        CycleGAN \cite{zhu2017unpaired} & 0.55 & 114.47 & 0.50 & 106.02 & 0.46 & 189.36 & 0.47 & \textbf{61.51} \\
    \hline
        ArtFlow \cite{an2021artflow} & 0.48 & \textbf{103.65} & 0.47 & 136.28 & 0.40 &148.80 & 0.35 & 144.03  \\
    \hline
        UNIT \cite{liu2017unsupervised} & 0.79 & 111.93 & 0.77 & 82.23 & 0.68 & 116.32 & 0.48 & 105.75 \\
    \hline
        Ours & \textbf{0.82} & 105.64 & \textbf{0.83} & \textbf{74.14} & \textbf{0.83} & \textbf{68.46}  & \textbf{0.71} & 70.27 \\
    \hline
    \end{tabular}}
    \label{tab:comparison}
\end{table*}
We also verify the performance of our model on the transform between CT and MRI. The datasets consist of the slices from the SynthRAD2023 Challenge \cite{adrian_thummerer_2023_7746020}, which contain several pairs of 3d CT and MR images for brains. For our datasets, we choose \textbf{170} pairs for training and \textbf{10} pairs for test for both CT and MR images, and for each 3d image, we choose 100 central slices. In this experiments, we also use vq-vae-2 to encode and decode both the CT and MR images. The sizes of CT and MR are both $(224, 168)$ and the sizes of the feature spaces after vq-vae-2 are $(128, 28, 21)$ and $(128, 56, 42)$ respectively. Then we trained our flow model for $1000$ epochs via optimizer AdamW with initial learning rate $10^{-3}$ and weight decay $10^{-5}$. The flow structure we use is the same as in the experiments of MRI T1/T2. The weight of the OT penalty is $10^{-3}$. As we can see from Tab. \ref{tab:comparison}, our proposed model performs the best with regard to both SSIM and FID scores compared to the other flow-based and GAN-based models. Some examples are shown in Fig.  \ref{fig:ct-mr} and the quantitative results on all the test images are also present in Tab.  \ref{tab:comparison}. As we can see, from  both the visual quality and quantitative indices (SSIM and FID)  comparison to the other models, our SyMOT-Flow achieves the best performances on both CT and MR image modalities.

\section{Conclusions} \label{Conclusion}
In this paper, we propose a novel symmetric flow model, named SyMOT-Flow, to learn a transformation between two unknown distributions from a set of samples drawn from each distribution, respectively. SyMOT-Flow leverages the maximum mean discrepancy (MMD) as a metric for comparing distributions. To enhance the generative performance of both forward and backward flows, a symmetrical design of the reversed component is incorporated based on the invertibility of the flow structure. Additionally, by treating the MMD as a relaxation of the equality constraint in the original optimal transport (OT) problem, SyMOT-Flow can also learn an asymptotic solution to OT. Besides, we provide some theoretical results regarding the feasibility of the proposed model and the connections to the OT solution. In the experimental section, SyMOT-Flow is evaluated on toy examples for illustration and real-world datasets, showcasing the generative samples and the transformation process achieved by the model. Furthermore, ablation studies are conducted to investigate the impact of the reversed flow constraint and the weight parameter on the OT cost. These experiments contribute to a better understanding of the effectiveness and robustness of SyMOT-Flow in practical scenarios.

\appendix
\section{Illustrative examples} 
Here are another examples of the comparison between the SWOT-Flow and our SyMOT-Flow model. The Example 3 is the result between two Gaussian distributions and Example 4 is the result between two groups of 8 Gaussians.
\begin{figure}[h]
	\begin{center}
	\scalebox{.9}[.9]{
		\begin{tabular}{c@{\vspace{-2pt}}c@{\vspace{-2pt}}c@{\vspace{-2pt}}c@{\vspace{-2pt}}c@{\hspace{1pt}}c@{\hspace{1pt}}c}
		   & Training Data & \multicolumn{1}{c}{SWOT-Flow}& \multicolumn{1}{c}{ MMD}&   \multicolumn{1}{c}{SyMOT-Flow}\\
			\put(-10,0){\rotatebox{90}{\textbf{Example 3}} }  &
			\includegraphics[width=.25\linewidth,height=.25\linewidth]{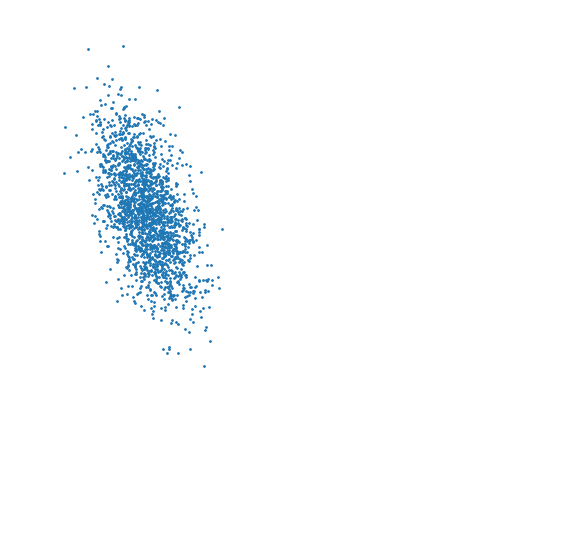}&	
			\includegraphics[width=.25\linewidth,height=.25\linewidth]{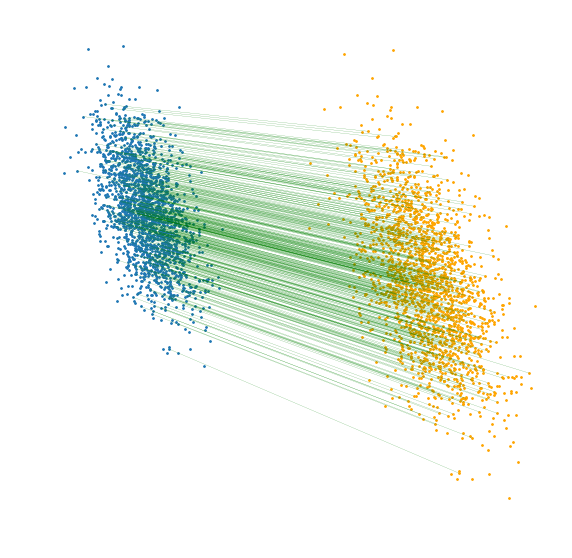}&
            \includegraphics[width=.25\linewidth,height=.25\linewidth]{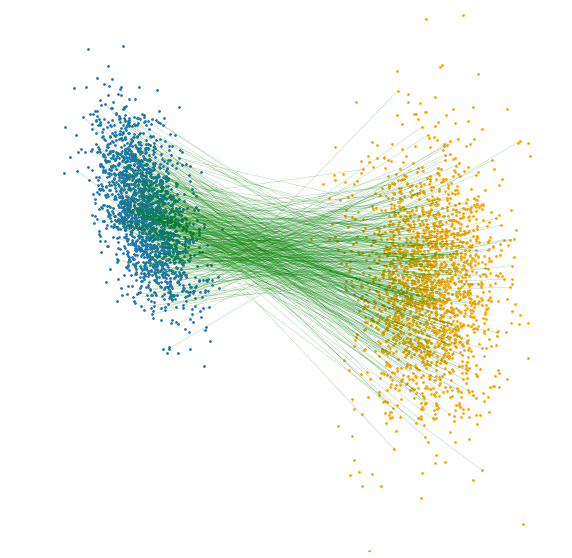}&
            \includegraphics[width=.25\linewidth,height=.25\linewidth]{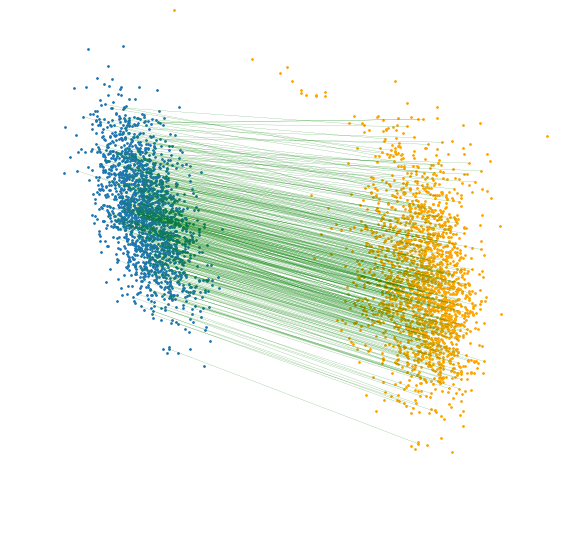}\\
           & \includegraphics[width=.25\linewidth,height=.25\linewidth]{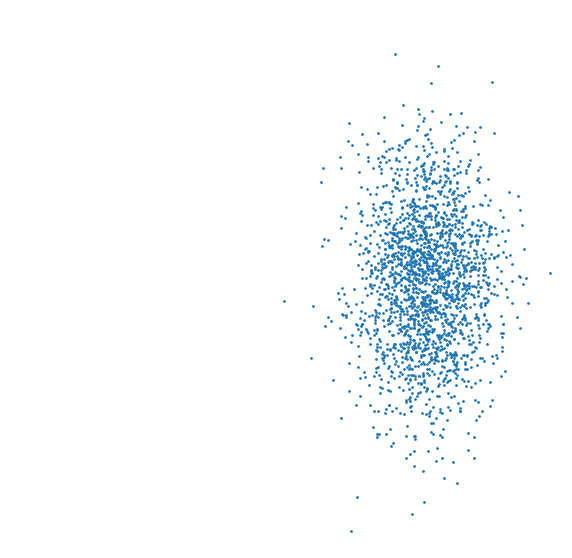}&	
             \includegraphics[width=.25\linewidth,height=.25\linewidth]{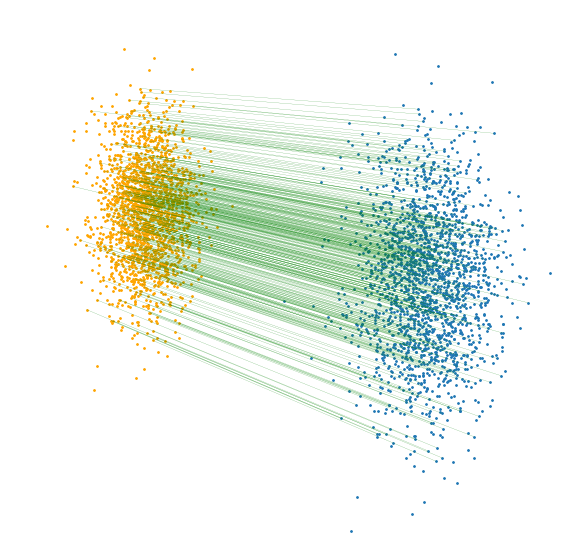}&
            \includegraphics[width=.25\linewidth,height=.25\linewidth]{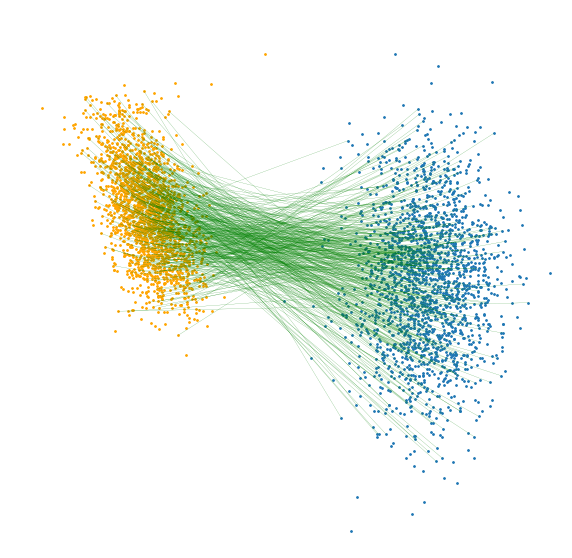}&
            \includegraphics[width=.25\linewidth,height=.25\linewidth]{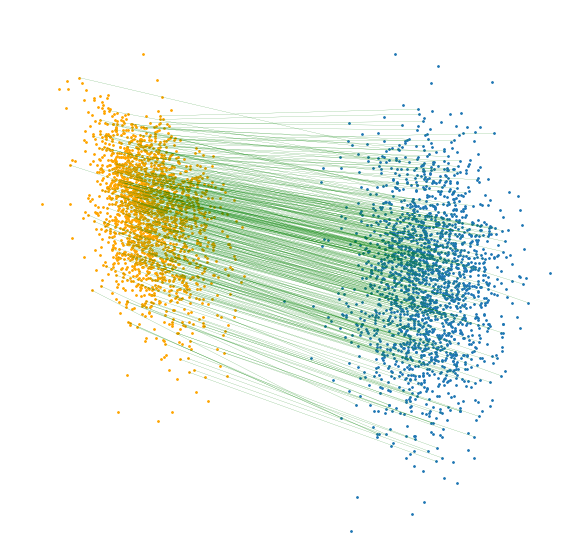}\\
        \put(-10,0){\rotatebox{90}{\textbf{Example 4}} } &
		\includegraphics[width=.25\linewidth,height=.25\linewidth]{images/8gauss28gauss/8g28g_8g1.png}&
			\includegraphics[width=.25\linewidth,height=.25\linewidth]{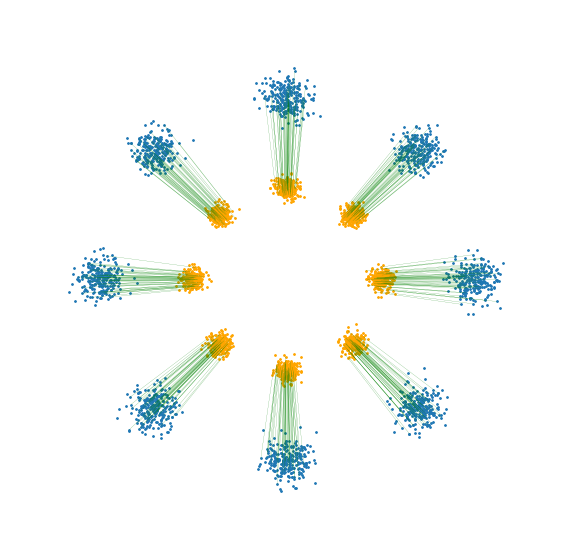}&
			\includegraphics[width=.25\linewidth,height=.25\linewidth]{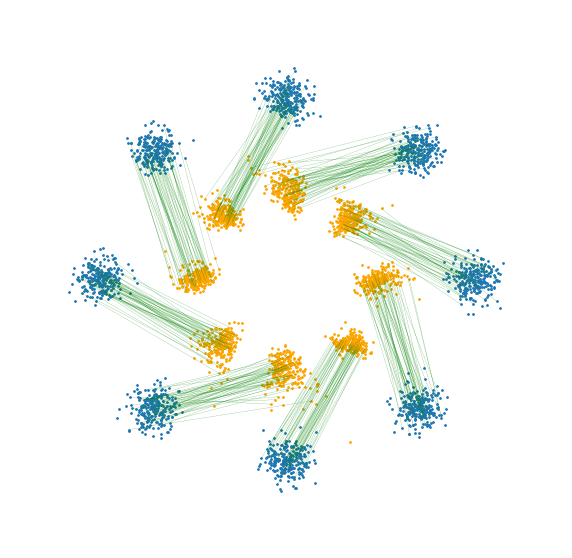}&
			\includegraphics[width=.25\linewidth,height=.25\linewidth]{images/8gauss28gauss/8g28g_o_1.png}\\
		&	\includegraphics[width=.25\linewidth,height=.25\linewidth]{images/8gauss28gauss/8g28g_8g2.png}&
			\includegraphics[width=.25\linewidth,height=.25\linewidth]{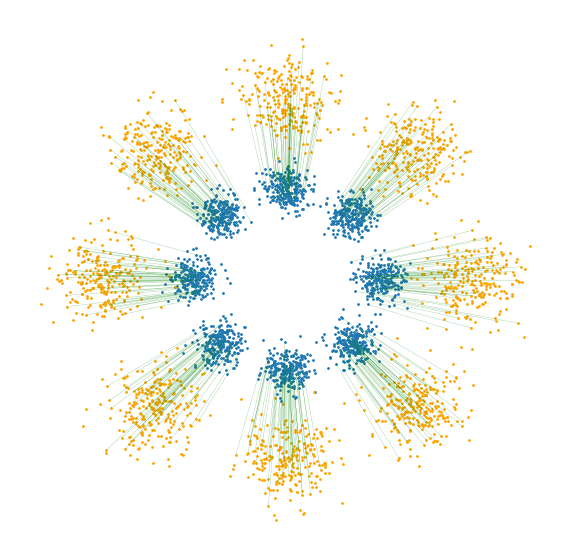}&
			\includegraphics[width=.25\linewidth,height=.25\linewidth]{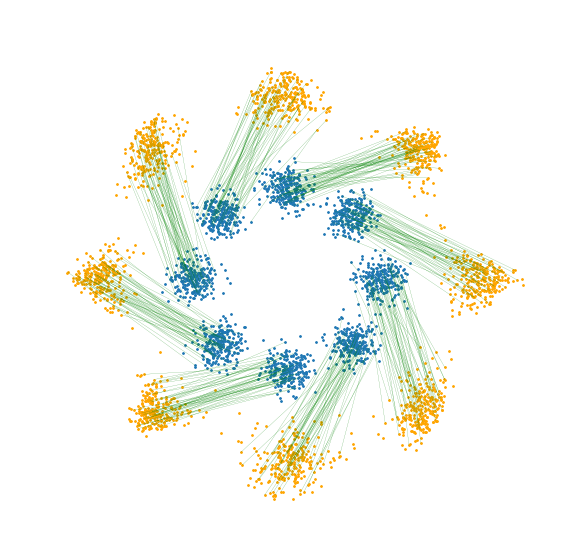}&
			\includegraphics[width=.25\linewidth,height=.25\linewidth]{images/8gauss28gauss/8g28g_o_2.png}
		\end{tabular}
		}
		\caption { The input (blue) and generated points (orange) by learned map (green lines) between two  sets of training data with different methods. Example 3 is between two Gauss distributions with different means and covariance and Example 4 is between two 8 Gauss distributions different means and covariance. }
	\end{center}
\end{figure}

\section{Ablation Study}
Here are some more ablation studies about the influence of the reverse cost on different datasets. The Example 1 is the result between Moons to Circles, Example 2 is the result between two Gaussian distributions and Example 3 is the result between two groups of linear Gaussians, which means the mean value of each Gauss lies in a line for each group. 

Moreover, Tabel \ref{tab:ablation} is the comparison of OT cost and MMD distance between the one direction flow and our SyMOT-Flow. As we can see that, compared to the SyMOT-Flow, the results of ene direction flow have larger OT cost and MMD distance on almost all the four datasets, which means that the reverse cost does work on the improvement to the flow model. Besides, from the visualization of Example 1 to 3, we can also find that the one direction flow has much worse performance than SyMOT-Flow, especially the generations of the backward direction.
\begin{figure}[htbp]
    \centering
    \scalebox{.9}[.9]{
        \begin{tabular}{c@{\vspace{-7pt}}c@{\vspace{-7pt}}c@{\vspace{-7pt}}c@{\vspace{-7pt}}c@{\hspace{1pt}}c@{\hspace{1pt}}c}
            Training Data & \multicolumn{1}{c}{One Direction}&   \multicolumn{1}{c}{SyMOT-Flow}\\
            \put(-10,0){\rotatebox{90}{\textbf{Example 1}} } 
            \includegraphics[width=.33\linewidth,height=.33\linewidth]{images/moon2circle/m2c_m.png}&	
            \includegraphics[width=.33\linewidth,height=.33\linewidth]{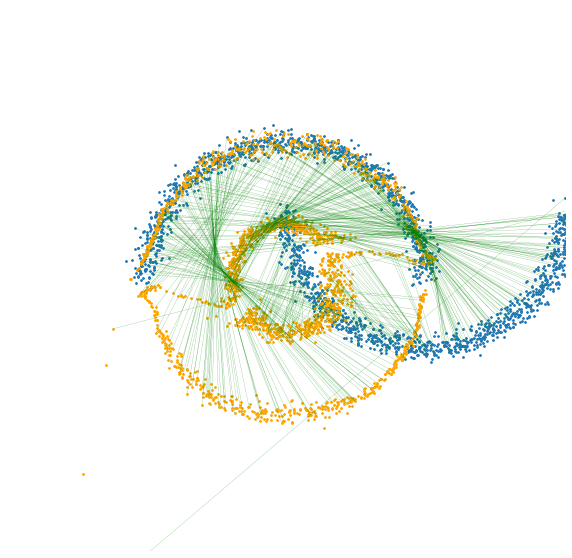}&
            \includegraphics[width=.33\linewidth,height=.33\linewidth]{images/moon2circle/m2c_o_1.png}\\
            \includegraphics[width=.33\linewidth,height=.33\linewidth]{images/moon2circle/m2c_c.png}&
            \includegraphics[width=.33\linewidth,height=.33\linewidth]{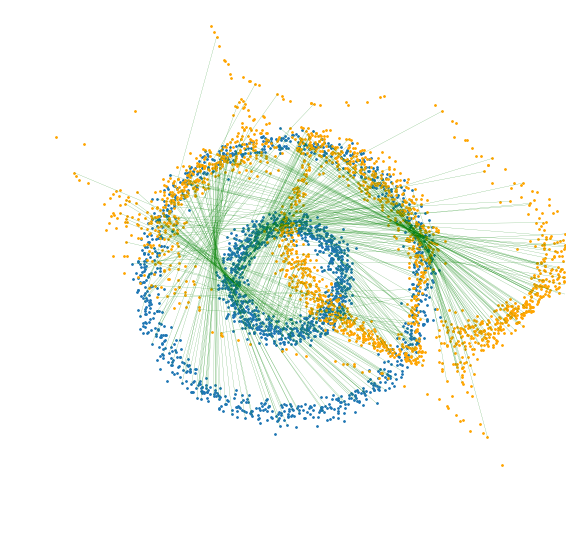}&	
            \includegraphics[width=.33\linewidth,height=.33\linewidth]{images/moon2circle/m2c_o_2.png}\\
            \put(-10,0){\rotatebox{90}{\textbf{Example 2}} } 
            \includegraphics[width=.33\linewidth,height=.33\linewidth]{images/gauss2gauss/g2g_g1.png}&	
            \includegraphics[width=.33\linewidth,height=.33\linewidth]{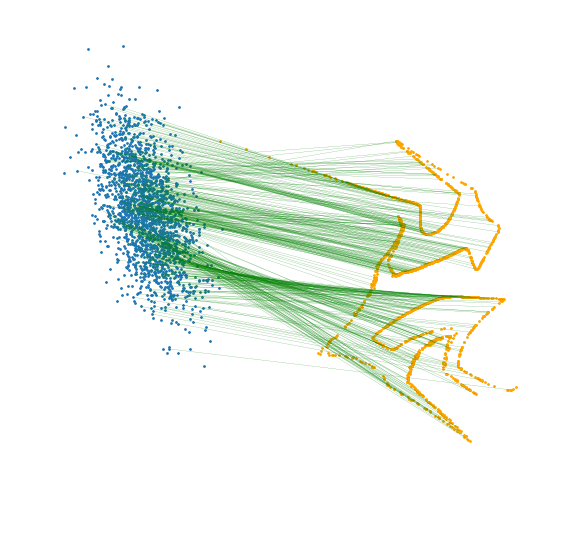}&
            \includegraphics[width=.33\linewidth,height=.33\linewidth]{images/gauss2gauss/g2g_o_1.png}\\
            \includegraphics[width=.33\linewidth,height=.33\linewidth]{images/gauss2gauss/g2g_g2.png}&
            \includegraphics[width=.33\linewidth,height=.33\linewidth]{images/ablation/g2g_1.png}&	
            \includegraphics[width=.33\linewidth,height=.33\linewidth]{images/gauss2gauss/g2g_o_2.png}\\
        \end{tabular}
        }
    \caption{Some Toy Examples}
\end{figure}

\begin{figure}[H]
	\begin{center}
	\scalebox{.9}[.9]{
		\begin{tabular}{c@{\vspace{-7pt}}c@{\vspace{-7pt}}c@{\vspace{-7pt}}c@{\vspace{-7pt}}c@{\hspace{1pt}}c@{\hspace{1pt}}c}
		    Training Data & \multicolumn{1}{c}{One Direction}&   \multicolumn{1}{c}{SyMOT-Flow}\\
		    \put(-10,0){\rotatebox{90}{\textbf{Example 3}} } 
			\includegraphics[width=.3\linewidth,height=.3\linewidth]{images/linegauss/lg_g1.png}&	
			\includegraphics[width=.33\linewidth,height=.33\linewidth]{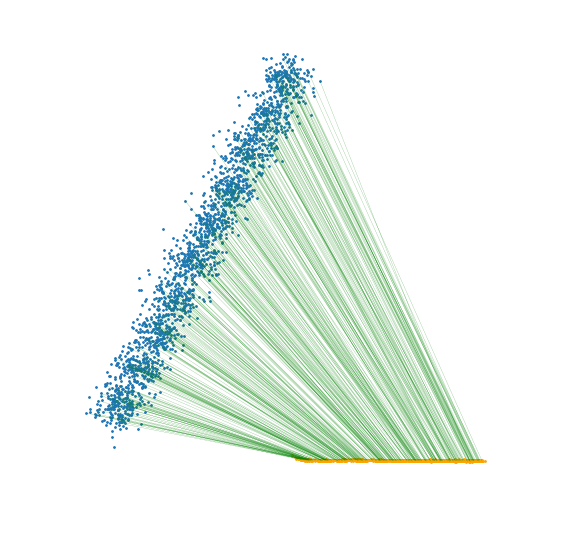}&
            \includegraphics[width=.33\linewidth,height=.33\linewidth]{images/linegauss/lg_o_1.png}\\
            \includegraphics[width=.33\linewidth,height=.33\linewidth]{images/linegauss/lg_g2.png}&
            \includegraphics[width=.33\linewidth,height=.33\linewidth]{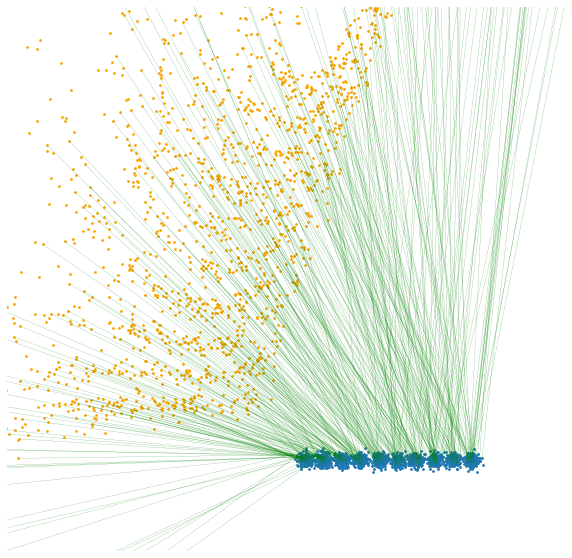}&	
            \includegraphics[width=.33\linewidth,height=.33\linewidth]{images/linegauss/lg_o_2.png}\\
		\end{tabular}
		}
	\end{center}
    \caption{Some Toy Examples}
\end{figure}

\begin{table}[H]
    \caption{The OT cost and MMD distance of forward/backward direction of flow with and without reverse cost.}
    \label{tab:ablation}
    \centering
    \begin{tabular}{lllll}
    \hline
      & Method  &     One Direction     & SyMOT-Flow \\
      \hline
    \multirow{4}{*}{\makecell[l]{OT\\ Cost}}&Moons to Circles & 0.465/0.715  &\textbf{0.295}/\textbf{0.288} \\
          & Gauss to Gauss & 33.301/1.9e7   & \textbf{32.459}/\textbf{32.475} \\
         & 8 Gauss to 8 Gauss & 40.333/41.441 & \textbf{7.539}/\textbf{7.483} \\
          &  Linear Gauss &  \textbf{138.939}/4.6e4  &139.780/\textbf{134.318} \\
          \hline
    \multirow{4}{*}{\makecell[l]{MMD\\ distance}} &    Moons to Circles & 1.3e-2/4.5e-2& \textbf{2.9e-3}/\textbf{2.7e-3} \\
        &Gauss to Gauss & \textbf{1.4e-2}/2.0e-1 & 6.6e-2/6.3e-2 \\
        &8 Gauss to 8 Gauss & 1.2e-2/2.9e-2 & \textbf{4.2e-3}/\textbf{2.5e-3} \\
        &Linear Gauss &  2.5e-3/6.6e-2  & \textbf{1.1e-3}/\textbf{1.3e-3} \\
    \hline
    \end{tabular}
\end{table}

\bibliographystyle{siamplain}
\bibliography{ref}

\end{document}